%% file: main.tex
\documentclass[lettersize,journal]{IEEEtran}
\usepackage{amsmath,amsfonts}
\usepackage{algorithmic}
\usepackage{algorithm}
\usepackage{array}
\usepackage[caption=false,font=normalsize,labelfont=sf,textfont=sf]{subfig}
\usepackage{textcomp}
\usepackage{stfloats}
\usepackage{url}
\usepackage{verbatim}
\usepackage{graphicx}
\usepackage{cite}
\usepackage{graphicx}  
\usepackage{caption}  
\usepackage{multirow}
\usepackage{mathabx}
\usepackage{amssymb}
\usepackage{pifont}
\usepackage[table,xcdraw]{xcolor}
\usepackage{lineno, hyperref}

\begin{document}
\newcommand{\gl}[1]{{\color{blue}(GL: {#1})}}  
\newcommand{\muyi}[1]{{\color{red}(muyi: {#1})}}  

\title{Vision Mamba in Remote Sensing: A Comprehensive Survey of Techniques, Applications and Outlook}

\author{
{Muyi Bao, Shuchang Lyu, Zhaoyang Xu, Huiyu Zhou, Jinchang Ren, \\ Shiming Xiang, Xiangtai Li, Guangliang Cheng$^\dagger$}
\thanks{Mr. Muyi Bao is with the School of Advanced Technology, Xi’an Jiaotong-Liverpool University.}
\thanks{Dr. Shuchang Lyu is with the School of Electronics and Information Engineering, Beihang University.} 
\thanks{Dr. Zhaoyang Xu is with the Department of Paediatrics, Cambridge University.} \
\thanks{Prof. Huiyu Zhou is with the School of Computing and Mathematical Sciences, University of Leicester.} 
\thanks{Prof. Jinchang Ren is with the Department of Computing Science, Robert Gordon University.} 
\thanks{Prof. Shiming Xiang is with the National Laboratory of Pattern Recognition, Institute of Automation, Chinese Academy of Sciences.} 
\thanks{Dr. Xiangtai Li is with Nanyang Technological University, Singapore.} 
\thanks{Dr. Guangliang Cheng is with the Department of Computer Science, University of Liverpool.} 
\thanks{$^\dagger$Corresponding author: Guangliang Cheng}
} 

\markboth{Journal of \LaTeX\ Class Files,~Vol.~14, No.~8, August~2021}%
{Shell \MakeLowercase{\textit{et al.}}: A Sample Article Using IEEEtran.cls for IEEE Journals}


\maketitle

\begin{abstract}
Deep learning has profoundly transformed remote sensing, yet prevailing architectures like Convolutional Neural Networks (CNNs) and Vision Transformers (ViTs) remain constrained by critical trade-offs: CNNs suffer from limited receptive fields, while ViTs grapple with quadratic computational complexity, hindering their scalability for high-resolution remote sensing data. State Space Models (SSMs), particularly the recently proposed Mamba architecture, have emerged as a paradigm-shifting solution, combining linear computational scaling with global context modeling. This survey presents a comprehensive review of Mamba-based methodologies in remote sensing, systematically analyzing about 120 Mamba-based remote sensing studies to construct a holistic taxonomy of innovations and applications. Our contributions are structured across five dimensions: (i) foundational principles of vision Mamba architectures, (ii) micro-architectural advancements such as adaptive scan strategies and hybrid SSM formulations, (iii) macro-architectural integrations, including CNN-Transformer-Mamba hybrids and frequency-domain adaptations, (iv) rigorous benchmarking against state-of-the-art methods in multiple application tasks, such as object detection, semantic segmentation, change detection, etc. and (v) critical analysis of unresolved challenges with actionable future directions. By bridging the gap between SSM theory and remote sensing practice, this survey establishes Mamba as a transformative framework for remote sensing analysis. 
To our knowledge, this paper is the first systematic review of Mamba architectures in remote sensing. Our work provides a structured foundation for advancing research in remote sensing systems through SSM-based methods. We curate an open-source repository (\url{https://github.com/BaoBao0926/Awesome-Mamba-in-Remote-Sensing}) to foster community-driven advancements.

\end{abstract}


\input{Sec/1_intro}

\input{Sec/2_preliminary}
\input{Sec/3_visionMamba}
\input{Sec/4_micro}

\input{Sec/5_macro}
\input{Sec/6_downstream}

\input{Sec/7_future}

\section{Conclusion} \label{sec:conclusion}
Mamba architectures have rapidly emerged as a promising alternative to conventional CNN-based and Transformer-based models in remote sensing applications, primarily owing to their linear computational complexity, dynamic feature selection via input-dependent parameterization, and efficient long-range dependency modeling capabilities. This survey systematically summarized the evolution of Mamba-based methods in remote sensing, beginning with a concise overview of vision Mamba backbone networks. Then, we conclude a systematic analysis of both micro-architectural advancement, including enhanced SSM formula, scan strategies and multi-modal and bi-temporal feature interaction, and macro-architectural developments encompassing hybrid CNN/Transformer integrations, framework substitutions in existing frameworks, learning paradigms, and frequency-domain operations. 
Some structured taxonomies were established to systematically organize these technological advancements, providing researchers with clear pathways for methodology comparison and selection. 
Furthermore, we identified critical challenges and proposed promising research directions, which holds substantial potential to advance Mamba's capabilities in remote sensing. 
These insights aim to catalyze future investigations and foster the development of next-generation remote sensing systems powered by Mamba-based models.

\section*{Acknowledgments}
This is acknowledgment

\bibliography{reference}
\bibliographystyle{IEEEtran}



\newpage

 




\vfill

\end{document}

%% file: Sec/1_intro.tex
\section{Introduction}

Recent advances in remote sensing have witnessed remarkable progress through deep learning methodologies for extracting features from complex data \cite{survey5, survey_remote}. 
Conventional architectures, particularly Convolutional Neural Networks (CNNs) and Vision Transformers (ViTs) \cite{ViT,Transformer}, have demonstrated notable success in this domain. However, the inherent characteristics of remote sensing imagery challenge the further enhancement of CNNs- and ViTs-based networks: (1) \textbf{Rich Spatial Dependencies}: Remote sensing images exhibit complex spatial relationships that often exceed the local modeling capabilities of CNNs with their limited receptive fields, despite linear computational complexity and versatile modeling capabilities \cite{1,2,3}; (2) \textbf{High Resolution}: The extremely high resolutions of remote sensing images impose prohibitive computational demands on Transformer-based models, often resulting in unacceptable levels of computational complexity. Despite these inherent constraints, CNNs and ViTs have long dominated remote sensing applications, prompting researchers to pursue architectures that can achieve global modeling with linear computational efficiency.

To address these challenges, state space models (SSMs)-based methods have emerged as a promising alternative, offering both linear computational complexity and global modeling capabilities. These two characteristics of SSMs precisely overcome the limitations of CNNs and ViTs in remote sensing applications.
Rooted in classical system theory, SSMs have found widespread applications across disciplines including reinforcement learning \cite{10}, computational neuroscience \cite{11}, and control systems \cite{kalmanfilter}. The SSM maps input sequences to latent states that encapsulate historical context, enabling sequential prediction according to the hidden state. Some work has tried to incorporate SSM into the deep learning framework \cite{s4, HIPPO, HIPPO2, H3, 14, 15, 16, 17, ssm, mamba, Mambav2}. Early SSM implementations faced computational bottlenecks until the Structured State Space sequence (S4) model \cite{s4} addressed these limitations through parameterized state matrices. Then, Mamba \cite{mamba} introduces Selective State Space for Sequences (S6) that incorporates dynamic time-aware mechanisms via regressed step size parameters, enabling context-aware information propagation or forgetting through hidden states. The Mamba architecture \cite{mamba} further advanced this paradigm through simplified gated SSM blocks, achieving state-of-the-art (SOTA) performance. 
With its linear scalability for high-resolution imagery, ability to capture long-range spatial dependencies and exceptional capability of feature representation, Mamba-based architectures have demonstrated strong potential as a next-generation solution for remote sensing tasks, bridging the gap between computational efficiency and global modeling.

Originally developed for natural language processing (NLP) with landmark studies \cite{mamba}, Mamba technology has swiftly expanded its application to computer vision (CV). Similar to the ViT \cite{ViT}, innovative Mamba-based architectures like Vision Mamba (Vim) \cite{VisionMamba} and Visual Mamba (VMamba) \cite{VMamba} employ patch embedding and the multi-directional scan strategy to convert 2D images into 1D sequences, making them compatible with Mamba's processing paradigm. 
Building on these foundational works, the remote sensing community has rapidly explore Mamba's potential, with numerous studies \cite{12-DMM, 32-RemoteDet-Mamba, 37-HSDet-Mamba, 1-RS-Mamba, 2-RS3Mamba, 3-RTMamba, DBLP:journals/lgrs/LiuCSTW25, 7-RSDehamba, 4-FreMamba, 17-RSMamba, 19-Mamba-CR,28-FusionMamba, 29-MSFMamba} adapting Mamba architecture to overcome domain-specific challenges. These innovations have propelled Mamba-based models to achieve the SOTA performance across various remote sensing tasks, including object detection \cite{12-DMM, 32-RemoteDet-Mamba, 37-HSDet-Mamba}, dense prediction \cite{1-RS-Mamba, 2-RS3Mamba, 3-RTMamba, DBLP:journals/lgrs/LiuCSTW25}, and others \cite{7-RSDehamba, 4-FreMamba, 17-RSMamba, 19-Mamba-CR,28-FusionMamba, 29-MSFMamba}.

\begin{figure*}[htbp]
\centering
\includegraphics[width=1\textwidth]{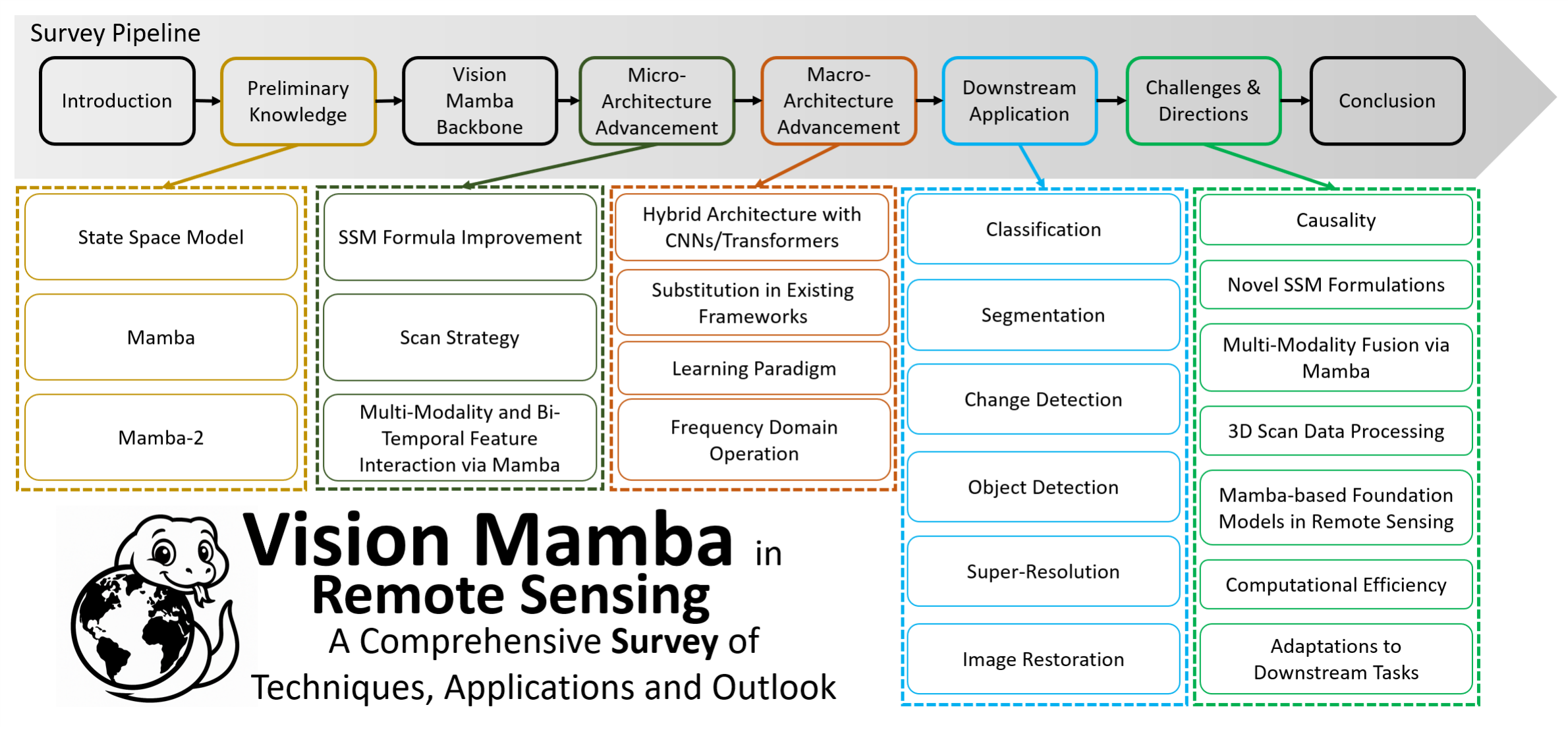}
\caption{A diagram that summarizes the pipeline of the survey.}
\label{fig:outline}
\end{figure*}

Although there are several surveys \cite{survey1, surveyon2, survey3, survey4, survey6, survey5}, they tend to emphasize the problem-solving approaches specific to natural imagery, often overlooking or underrepresenting the distinctive features of remote sensing images. 
Consequently, we have conducted a comprehensive survey dedicated to the remote sensing domain, detailing the current research progress of Mamba technology in this field, its applications, and some potential future trends. 
To the best of our knowledge, this represents the \textbf{first} survey in the remote sensing field with Vision Mamba technology, which would further benefit the remote sensing community.
%


\par \textbf{Contribution: } This survey makes several key contributions to the emerging field of Mamba-based architectures in remote sensing. First, we provide a systematic introduction to Mamba's foundational concepts, followed by a brief review of approximately 20 vision Mamba backbone papers. Then, we review and synthesize about 120 remote sensing papers. Our analysis adopts both micro- and macro-architectural perspectives to offer a holistic understanding of recent Mamba advancements in the remote sensing domain.

\textbf{Micro-architecture Advancements:} We examine three critical advancements within the inner mechanism of Mamba blocks:

\begin{itemize}
     \item SSM Formula Enhancement: While Mamba~\cite{mamba} established the initial adaptation of SSMs for deep learning, subsequent works have further refined this formulation. To our knowledge, we are the first to systematically review SSM formula improvements in this domain.

    \item Scan Strategy: We propose a novel taxonomy for scan strategies (crucial for processing 2D/3D imagery as 1D sequences) comprising five elements: \textit{preprocessing, scan sampling, scan direction, scan pattern,} and \textit{postprocessing}. This framework provides the first comprehensive treatment that includes preprocessing and postprocessing stages into the scan strategy.
    
    \item Multi-modal and Bi-temporal Feature Interaction: Focusing on remote sensing applications, we categorize existing approaches into four distinct methodologies for handling multi-modal data and bi-temporal interactions. To our knowledge, our analysis offers the first detailed examination of the multi-modal and bi-temporal interaction based on the Mamba architectures.
\end{itemize}

\textbf{Macro-architecture Advancement:} We analyze four innovative advancements for overall architecture advancement:

\begin{itemize}
    \item Hybrid Architectures: We analyze integration for combining Mamba with CNNs and Transformers, examining both basic stackable blocks and overall architectural designs.
    
    \item Substitution in Framework Adaptation: We survey the incorporation of Mamba blocks into established frameworks, including U-Net~\cite{UNet}, You Only Look Once (YOLO)~\cite{YOLO} and Diffusion models~\cite{diffusion}.
    
    \item Learning Paradigms: In addition to the conventional supervised learning, our review encompasses applications of unsupervised, self-supervised, and prompt learning paradigms.
    
    \item Frequency Domain Operations: We document implementations incorporating frequency domain operations within Mamba architecture, including Fast Fourier Transform, 2D discrete cosine transform and wavelet transform.
\end{itemize}

Then, we conducted an exhaustive summary and analysis of Mamba's applications in remote sensing imagery, such as classification, object detection and semantic segmentation. This involved collating and comparing the performance of modern architectural approaches, including Mamba, Transformers, and CNNs, facilitating a comprehensive comparative study that highlights the strengths and weaknesses of each architecture within this specific application area. Finally, drawing on the current developmental trends of Mamba in remote sensing, we proposed several potential future trends. These suggestions aim to provide valuable insights and serve as a reference for ongoing and future research efforts, potentially guiding the next steps in the evolution of remote sensing technologies.

\par \textbf{Survey Scope:} This survey comprehensively examines the foundational literature related to Vision Mamba within the field of remote sensing. We restricted our review to works that were published or appeared as preprints on Arxiv prior to February 2025. Although there are many preprints or published works with vision mamba in natural images and videos, we only include the most representative works. 

\par \textbf{Organization:} The rest of the survey is organized as follows. Overall, Fig. \ref{fig:outline} shows the pipeline of this survey. We first introduce the preliminary knowledge of Mamba in Section~\ref{sec:PreKnow}. This is followed by a detailed review of the backbone architectures of Vision Mamba in Section~\ref{sec:vision_mamba}. Then, we delve into the specific advancements of Mamba architecture, discussing micro-architecture advancements in Section~\ref{sec:micro} and macro-architecture advancements in Section~\ref{sec:macro}. We compare the experiment results of several downstream tasks in Section~\ref{sec:downstramtask}. Finally, we raise the current challenges and future directions in Section~\ref{sec:direction} and conclude the survey in Section~\ref{sec:conclusion}.

%% file: Sec/2_preliminary.tex
\section{Preliminary Knowledge} \label{sec:PreKnow}

\subsection{State Space Model}

Based on the Kalman filter \cite{kalmanfilter}, State Space Models (SSMs) \cite{ssm} input a one-dimensional continuous sequence \(x(t) \in \mathbb{R}^{N}\) and convert \(x(t)\) and intermediate hidden state of last moment \(h(t) \in \mathbb{R}^N\) into hidden states of this moment \(h'(t) \in \mathbb{R}^N\) according to \(x(t)\) and the previous hidden state. The output \(y(t) \in \mathbb{R}\) is then computed based on \(x(t)\) and the hidden state of this moment \(h(t)\). This process can be formulated as follows:

\begin{align}
h'(t) &= \mathbf{A} h(t) + \mathbf{B} x(t) \label{eq:01}\\
y(t) &= \mathbf{C} h(t) + \mathbf{D} x(t) \label{eq:02}
\end{align}
, where \( \mathbf{A} \in \mathbb{R}^{N \times N} \) is a learnable state matrix; \( \mathbf{B} \in \mathbb{R}^{N \times 1} \), \( \mathbf{C} \in \mathbb{R}^{R \times 1} \), and \( \mathbf{D} \in \mathbb{R}^{R \times 1} \) are learnable projection parameters. In Eq. (\ref{eq:02}), \( \mathbf{D} x(t) \) is sometimes regarded as a residual connection in deep learning neural networks, and thus may be omitted. 

To handle the discrete input sequence \( \mathbf{x} = (x_0, x_1, \dots, x_L) \in \mathbb{R}^L \), the Structured State Space Sequence (S4) model \cite{s4} discretizes all learnable parameters in Eq. (\ref{eq:01}) and Eq. (\ref{eq:02}) by introducing a time step size \( \boldsymbol{\Delta} \), which is projected through a simple multi-layer perception (MLP). The time step size \( \boldsymbol{\Delta} \) can be interpreted as the resolution of the continuous input. Then, by applying the Zero-Order Hold (ZOH) method, the continuous parameters \( \mathbf{A} \) and \( \mathbf{B} \) are converted into their discrete counterparts \( \overline{\mathbf{A}} \), \( \overline{\mathbf{B}} \), and \( \overline{\mathbf{C}} \) as follows:

\begin{align}
\overline{\mathbf{A}} &= \text{exp}(\boldsymbol{\Delta}\mathbf{A}) \label{eq:03}\\
\overline{\mathbf{B}} &= (\boldsymbol{\Delta}\mathbf{A})^{-1}( \text{exp}(\boldsymbol{\Delta}\mathbf{A})-\mathbf{I})\cdot\boldsymbol{\Delta}\mathbf{B}\approx\boldsymbol{\Delta}\mathbf{B} \label{eq:04}\\
\overline{\mathbf{C}} &= \mathbf{C} \label{eq:05}
\end{align}
, where \( \overline{\mathbf{A}} \in \mathbb{R}^{N \times N} \), \( \overline{\mathbf{B}} \in \mathbb{R}^{D \times N} \) and \( \overline{\mathbf{C}} \in \mathbb{R}^{D \times N} \). After discretization, Eq.~(\ref{eq:01}) and Eq.~(\ref{eq:02}) can be reformulated as follows:

\begin{align}
h_t &= \overline{\mathbf{A}} h_t + \overline{\mathbf{B}} x_t \label{eq:06}\\
y_t &= \overline{\mathbf{C}} h_t \label{eq:07}
\end{align}

This linear recurrence calculation can be accelerated by SSM convolution kernel as follows:

\begin{align}
y &= x \ast \overline{\mathbf{K}} \notag \\
\text{with} \quad \overline{\mathbf{K}} &= (\overline{\mathbf{CB}}, \overline{\mathbf{CAB}}, \dots, \overline{\mathbf{CA^{L-1}B}})
\end{align}
, where $\ast$ is the convolution operation and $\overline{\mathbf{K}} \in \mathbb{R}^{L}$ is the SSM kernel.

\subsection{Mamba: Selective SSM}

In the SSMs of S4, all learnable parameters remain fixed after training and do not dynamically change with the input. This linear time-invariant (LTI) nature is pointed out by Mamba \cite{mamba} as a fundamental limitation of SSM in the context-dependent reasoning. To address this limitation, Mamba introduces a selection mechanism. Specifically, the parameters \( \mathbf{B} \in \mathbb{R}^{B \times L \times N} \), \( \mathbf{C}  \in \mathbb{R}^{B \times L \times N}\), and \( \boldsymbol{\Delta}  \in \mathbb{R}^{B \times L \times N}\) in Eq. (\ref{eq:06}) and Eq. (\ref{eq:07}) are generated by projecting the input  \( \mathbf{x} \in \mathbb{R}^{B \times L \times N} \) through a simple learnable MLP, which enables input-dependent parameter values. The process can be formulated as follows:

\begin{align}
\mathbf{B}, \mathbf{C}, \mathbf{\Delta} = Linear(\textbf{x}) \label{eq:9}
\end{align}

\begin{figure*}[htbp]
\centering
\includegraphics[width=1\textwidth]{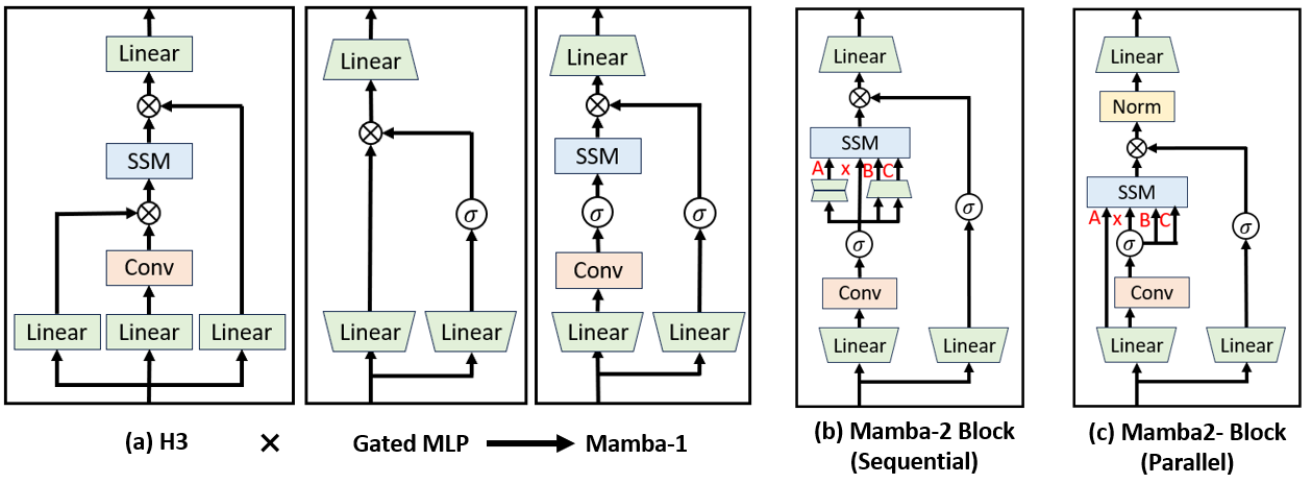}
\caption{The architecture of Mamba-1 \cite{mamba} and Mamba-2 \cite{Mambav2}. 
(a) Mamba-1, a simplified attention block that integrates the SSM transformation directly into the primary computational pathway of the MLP block and uses an activation function on the gating pathway.
(b) Sequential Mamba-2 Block, where operations are applied in series, and (c) Parallel Mamba-2 Block, where selected operations execute concurrently before fusion.
}
\label{fig:Mamba}
\end{figure*}

The core building block of Mamba is a simplified attention architecture. Unlike conventional SSM designs that employ a stacked combination of linear attention-like blocks and MLP blocks, similar to Transformer \cite{Transformer} architectures, Mamba unifies these two fundamental components into a single structure, forming the Mamba block. As depicted in Fig. \ref{fig:Mamba}.a, the Mamba block can be analyzed from two distinct perspectives. First, it substitutes the multiplicative gating mechanism in linear attention-like or H3 \cite{H3} blocks with an activation function. Second, it integrates the SSM transformation directly into the primary computational pathway of the MLP block. The overall architecture of Mamba is composed of multiple Mamba blocks, interspersed with standard normalization layers and residual connections. In detail, the activation function uses SiLU \cite{GeLUs} and Swish activation \cite{Swish}, and we normally call the activation score in the gating pathway as $z$. This architecture effectively captures long-range dependencies while maintaining the linear scalability characteristic of SSMs with respect to sequence length, becoming a new promising foundation model for CV.

\subsection{Mamba-2}
Mamba-2 introduces a unified architecture for sequence modeling through \textbf{Structured State Space Duality (SSD)}, bridging structured state space models (SSMs) and self-attention mechanisms \cite{Mambav2}. At its core, SSD reinterprets SSM computations as matrix multiplications over \textit{semiseparable matrices}—structured matrices with subquadratic parameterization and linear-time algorithms. This dual view enables two computation modes:  
\begin{itemize}
    \item A \textbf{linear-time recurrent mode} using scalar-identity transitions ($A_t = a_t I$), optimized via tensor contractions;
    \item A \textbf{quadratic attention-like mode} with data-dependent semiseparable masks ($L_{ij} = \prod_{k=j+1}^i a_k$), replacing softmax with cumulative gating \cite{katharopoulos2020transformers}.
\end{itemize}

\subsubsection{Architectural Innovations}
\label{sec:arch}
Key advancements include:  
1) \textbf{Parallel Parameterization}: Co-projecting SSM parameters ($A, B, C$) and values ($X$) akin to attention's $QKV$ mappings, reducing synchronization bottlenecks by 50\% compared to Mamba-1 \cite{mamba};  
2) \textbf{Multi-Input SSM (MIS) Heads}: Sharing $B/C$ projections across channels while keeping $A$ head-specific, enabling 8$\times$ larger state dimensions (e.g., $N=512$) with minimal overhead;  
3) \textbf{Hybrid Blocks}: As illustrated in Fig.~\ref{fig:Mamba}.b and Fig.~\ref{fig:Mamba}.c, Mamba-2 supports both \textit{sequential} and \textit{parallel} configurations of its core components (e.g., convolutional layers, SSM transformations, and nonlinear activations), allowing flexible trade-offs between computational efficiency and model capacity.


\subsubsection{Efficiency and Performance}
Mamba-2 achieves Pareto dominance over Transformers and Mamba-1, attaining \textbf{6.09 perplexity} on the Pile (2.7B parameters) with 3$\times$ faster training than FlashAttention-2 at 16K context \cite{FlashAttention, FlashAttention2}. Its blockwise SSD algorithm splits sequences into chunks, combining intra-chunk matrix multiplications and inter-chunk state transitions for 2–8$\times$ speedups. System optimizations include tensor parallelism with a single all-reduce per block and sequence-parallel state propagation for billion-token sequences. Evaluations highlight 98\% accuracy on memory-intensive tasks (e.g., multi-query associative recall), outperforming Mamba-1 by 16\% \cite{Griffin}.

%% file: Sec/3_visionMamba.tex
\section{Vision Mamba Backbones} \label{sec:vision_mamba}

In this section, we provide an overview of Vision Mamba backbones. For an in-depth exploration, readers are encouraged to refer to other related surveys \cite{survey1, surveyon2, survey6, survey5}. We begin by introducing two pioneering studies that first integrate Mamba into the computer vision domain, highlighting their core ideas, and then proceed to discuss related subsequent advancements.

\textbf{Vision Mamba (Vim)} \cite{VisionMamba} is the pioneering work that introduces Mamba to the field of computer vision. Inspired by the ViT framework \cite{ViT}, Vim first segments images into 2D patches, which are subsequently vectorized using CNNs. These vectorized patches are enhanced with positional embeddings to preserve spatial context, establishing a standard procedure adopted by subsequent studies. Additionally, Vim appends a class token in order to conduct classification. To overcome the inherent causality limitations of SSMs, Vim introduces a bidirectional scan strategy. Specifically, it first flattens the 2D patches into a 1D sequence and then employs both forward and backward pathways, scanning the sequence from the beginning to the end and from the end to the beginning, respectively. This non-hierarchical architecture of Vim comprises multiple identical Vim blocks.


\textbf{Visual Mamba (VMamba)} \cite{VMamba} addresses two critical challenges in applying SSMs to 2D images: 1) the inadequacy of using 1D CNNs before SSMs for capturing 2D spatial structures; 2) the intrinsic causal nature of SSMs, which converts 2D image patches into sequential 1D data, losing spatial information. To overcome these issues, VMamba introduces the Cross-Scan Module (CSM). Specifically, CSM replaces traditional 1D CNNs with 2D Depth-Wise CNN layers (DWConv). This adaptation better preserves and extracts spatial contextual information within images and has subsequently become a standard approach in subsequent researches. Additionally, CSM incorporates a 2D-Selective-Scan (SS2D) mechanism that systematically scans the entire image from four distinct directions: vertically and horizontally from both the top-left and bottom-right patches. It then aggregates the resulting four 1D sequences into a unified representation. This multi-directional scan strategy effectively mitigates the mismatch between the causal nature of selective SSMs and the spatial characteristics of images, enhancing the overall representation capability.


Considering one fundamental challenge in adapting SSMs to vision tasks lies in effectively transforming 2D spatial relationships into sequential representations, numerous works have focused on modifying scan strategies to increase the model's understanding of non-causal 2D data.
Building upon the approaches of Vim and VMamba, some work focuses on scan strategies (discussed in Section \ref{sec:scan pattern}). 
\textbf{Mamba-ND} \cite{MambaND} generalizes scan strategies to accommodate 3D data by employing block-level alternation of scan directions along the height, width, and time axes. 
\textbf{LocalVMamba} \cite{LocalMamba} integrates local window-based scanning (Fig.~\ref{fig:scan_4_pattern}.D) with learnable pathway weights through DARTS-inspired architecture search \cite{DARTS}.
\textbf{PlainMamba} \cite{plainmamba} and \textbf{FractalMamba} \cite{FractralMamba} employ a continuous 2D scan pattern (Fig.~\ref{fig:scan_4_pattern}.B) and Hilbert curve pattern (Fig.~\ref{fig:scan_4_pattern}.E), respectively, aiming to enhance contextual understanding capabilities and address the issue of discontinuity. 
Subsequently, some researchers have leveraged preprocessing methods (discussed in Section \ref{sec:mulplePre}) to enhance the performance of SSMs. \textbf{MS-VMamba} \cite{MSVMamba} applies multi-scale SSMs, while \textbf{GroupMamba} \cite{GroupMamba} divides the input feature into four groups in channel dimensions and applies a separate SSM independently to each group.

Some researchers have developed Mamba-based architecture on designs that are proven effective in CNN/Transformer blocks.
\textbf{MambaMixer} \cite{MambaMixer} follows the core idea of MLP-Mixer \cite{MLP-Mixer} to combine Mamba-based token mixing with channel mixing, and follows the design of DenseNet \cite{DenseNet} and DenseFormer \cite{DenseFormer} to allow blocks to access the previous features.
Following \cite{TransformerR}, \textbf{MambaR} \cite{MambaR} evenly inserts register tokens into the input sequences and recycles registers for final decision predictions, which can help the model focus on more semantic regions.
\textbf{ARM} \cite{ARM} follows the GPT families \cite{iGPT, GPT4, ChatGPT}, which utilizes the autoregressive modeling self-supervised framework. 
Similar to EViT \cite{EViT}, \textbf{MambaPruning} \cite{MambaTokenPruning} introduces a token pruning method specifically designed for Mamba-based models. 
Inspired by \cite{shuffleTrans}, \textbf{ShuffleMamba} \cite{shuffleMamba} introduces a training regularization technique that randomly shuffles token sequences during training to improve positional transformation invariance and overall model performance.

On one hand, some papers focus on the micro-architecture designs to enhance the model's ability.
\textbf{VSSD} \cite{VSSD} introduces non-causal processing for image data by decoupling interaction magnitudes between hidden states and tokens.
\textbf{SparX-Mamba} \cite{SparX} introduces an efficient sparse cross-layer feature aggregation method specifically designed for SSMs.
\textbf{Vim-F} \cite{Vim-F} integrates frequency domain information into the basic block to achieve a better global receptive field.
On the other hand, some work focuses on the macro-architecture advancement.
\textbf{SiMBA} \cite{SiMBA} utilizes hybrid sequence-channel modeling, effectively resolving the instability issues in large-size vision Mamba.
\textbf{StableMamba} \cite{StableMamba} establishes a robust interleaved Mamba-Attention framework that successfully scales SSMs to over 100 million parameters without resorting to knowledge distillation techniques. 
\textbf{MambaVision} \cite{MambaVision} redesigns the Mamba architecture and integrates CNNs and ViT blocks on the architecture level, aiming at enhancing modeling capability.

%% file: Sec/4_micro.tex
\section{Micro-Architecture Advancement} \label{sec:micro}

This section focuses on the inner mechanism within the micro-architecture of Mamba block. Three core points are summarized: (1) SSM Formula Improvement (Section \ref{sec:SSM}); (2) Scan Strategy (Section \ref{sec:scanstrategy}) and (3) Multi-Modal and Bi-Temporal Feature Interaction via Mamba (Section \ref{sec:multimodality}).

\subsection{SSM Formula Improvement} \label{sec:SSM}

\begin{figure*}[h!]
\centering
\includegraphics[width=1\textwidth]{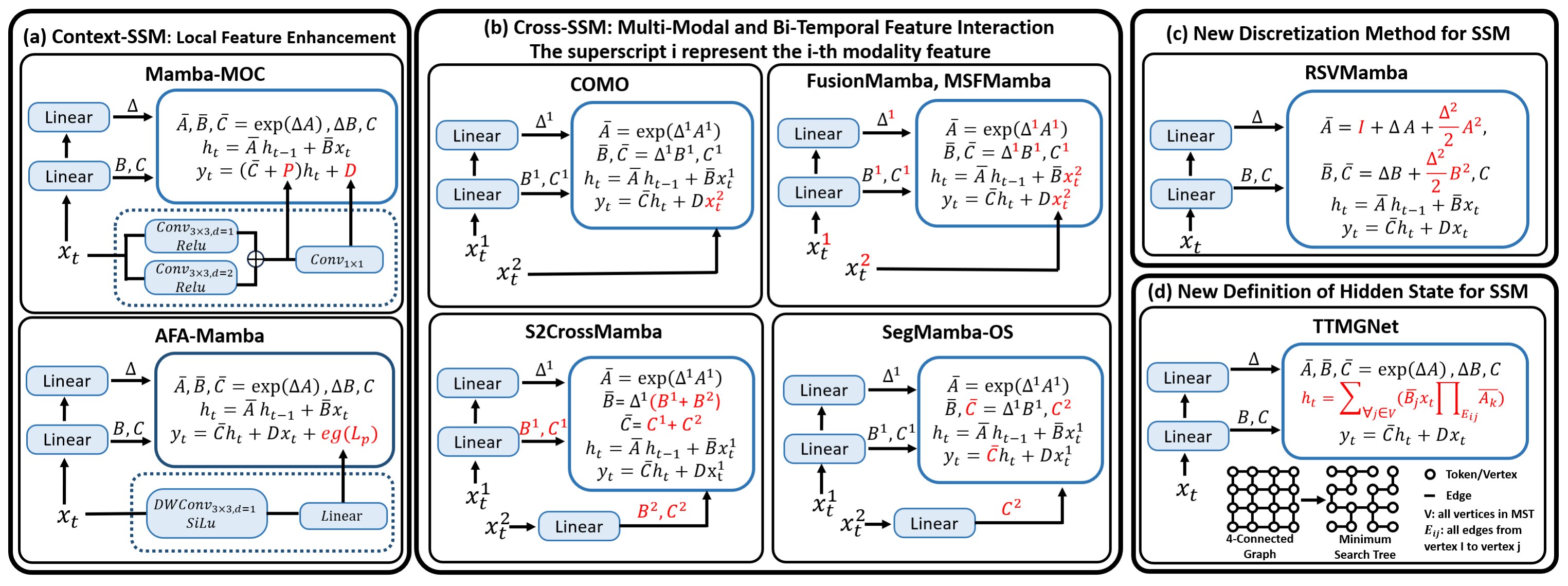}
\caption{Overview of Recent Advancements in SSM Formulations. This figure categorizes improvements into four major aspects: (a) Context-SSM \cite{62-Mamba-MOC, 35-AFA-Mamba} for local feature enhancement, integrating local context through CNNs; (b) Cross-SSM \cite{84-COMO, 28-FusionMamba, 29-MSFMamba, 5-S2CrossMamba, 47-SegMamba-OS} for multi-modal and bi-temporal feature interaction, enabling cross-modal/temporal feature interaction at the parameter level; (c) New discretization methods for SSM \cite{112-RSVMamba}, introducing second-order Runge-Kutta (RK2) for improved continuous-time approximation; and (d) Alternative hidden state definitions \cite{68-TTMGNet}, utilizing a Minimum Search Tree (MST) to redefine hidden state transitions. These refinements collectively enhance the efficiency and expressiveness of SSMs in various applications.}
\label{fig:SSM}
\end{figure*}

Vanilla Mamba initially adopts the ZOH rule to discretize continuous-time SSM and utilizes a selection mechanism to dynamically adjust model parameters based on the input sequence. Building upon this, several subsequent studies have modified the basic SSM formula to achieve enhanced performance and fulfill specific objectives. In total, 9 papers have contributed to the refinement of the SSM formula, which can be categorized into enhancing local context understanding \cite{62-Mamba-MOC, 35-AFA-Mamba}, interacting multi-modal and bi-temporal features \cite{5-S2CrossMamba,84-COMO, 28-FusionMamba, 29-MSFMamba, 47-SegMamba-OS}, utilizing alternative discretization methods for continuous signals \cite{112-RSVMamba} and using alternative definition of hidden states \cite{68-TTMGNet}. These contributions are visually summarized in Fig. \ref{fig:SSM}.

\subsubsection{\textbf{Context-SSM for Local Feature Enhancement}}
Context-SSM aims to naturally incorporate local context information into modeling long-range dependencies.
Mamba-MOCO \cite{62-Mamba-MOC}, building upon insights from MambaIRv2 \cite{MambaIRv2}, identifies two critical insights for improving SSMs in CV: (1) the output matrix $C$ in Eq. \ref{eq:07} functions as a "Query" for local context information, which can address the inherent limitation of transforming 2D images into 1D sequences; (2) enhancing local context allows the model to better focus on local interactions while preserving its ability to model global dependencies. In response to these insights, Mamba-MOC \cite{62-Mamba-MOC} introduces the Context-SSM, as depicted in Fig. \ref{fig:SSM}.a. Context-SSM incorporates two new parameters, $P$ and $D$, which are generated by multi-scale CNNs and a $1 \times 1$ CNN. The parameters $P$ and $D$ capture local context information, ensuring that the model can effectively capture both local context and long-range dependencies for the input data. Similarly, AFA-Mamba \cite{35-AFA-Mamba} simply adds local context features $eg(L_p)$ with the output equation together, which improves the ability to understand local context.

\subsubsection{\textbf{Cross-SSM for Multi-Modal and Bi-temporal Feature Interaction}} \label{sec:crossSSM}

In multimodal and bi-temporal settings, enabling interactions between features from different modalities is crucial for obtaining comprehensive representations. Recent advancements have introduced modifications to the SSM formula to inherently support cross-modal interactions, as highlighted in \cite{5-S2CrossMamba, 84-COMO, 28-FusionMamba, 29-MSFMamba, 47-SegMamba-OS} and illustrated in Fig.~\ref{fig:SSM}.b.

The COMO approach \cite{84-COMO} adopts an intuitive design, generating outputs by integrating hidden states from one modality with feature representations from another. This design facilitates dynamic and interactive information fusion between modalities. In contrast, S2CrossMamba \cite{5-S2CrossMamba} introduces an alternative mechanism by combining the parameters $B$ and $C$ derived from each modality to form combined parameters $\overline{B}$ and $\overline{C}$. This approach enables modality interaction directly at the parameter level of the SSM. Consequently, the resulting output incorporates feature information from both modalities, differing from approaches that initially combine modality features before projecting them into the state matrix.

Furthermore, FusionMamba \cite{28-FusionMamba}, MSFMamba \cite{29-MSFMamba}, and SegMamba-OS \cite{47-SegMamba-OS} share the objective of leveraging SSM parameters (such as $A$, $B$, $C$, and $\Delta$) from one modality to influence another modality’s feature representations. Specifically, SegMamba-OS \cite{47-SegMamba-OS} applies only the state matrix $C$ of the second modality directly to the output equation of the first modality's features. This strategy effectively narrows the semantic gap between modalities, enhancing cross-modal relevance.
Expanding upon this, FusionMamba \cite{28-FusionMamba} and MSFMamba \cite{29-MSFMamba} utilize all SSM parameters ($A$, $B$, $C$, and $\Delta$) from one modality to impact the feature representations of another modality. By influencing both the output equations and the hidden state updating processes, these methods comprehensively integrate cross-modal information, significantly strengthening the interactions between modalities.

\subsubsection{\textbf{New Discretization Method for SSM}}
Vanilla Mamba employs the ZOH rule to discretize continuous data, which serves as a fundamental approach for integrating SSMs within deep learning frameworks. Despite its widespread adoption, the ZOH rule exhibits a notable limitation, as highlighted by RSVMamba \cite{112-RSVMamba}: it approximates hidden states solely at discrete sampling points, overlooking dynamic variations occurring within intervals between these points. Consequently, this approximation can lead to local errors that accumulate as the frequency of sampling points increases.

To address this limitation, RSVMamba introduces the second-order Runge-Kutta (RK2) method \cite{RK2} for discretization, as illustrated in Fig. \ref{fig:SSM}.c. The RK2 approach enhances accuracy by computing intermediate states within each discrete interval, providing a more precise approximation of the underlying continuous-time dynamics. Moreover, the RK2 method maintains manageable computational complexity compared to higher-order alternatives, such as the fourth-order Runge-Kutta method, thus achieving an effective balance between accuracy and computational efficiency.

\subsubsection{\textbf{New Definition of the Hidden States}}
Beyond adjusting the SSM discretization, another promising direction involves redefining the hidden states themselves. TTMGNet \cite{68-TTMGNet} exemplifies this by first converting 2D features into a Minimum Spanning Tree (MST) based on cosine similarity. In this model, each token’s hidden state is associated with the state transmission matrices of neighboring tokens rather than being sequentially iterated through Eq. \ref{eq:06}. This conceptual modification is illustrated in Fig. \ref{fig:SSM}.d. Although tailored specifically to the tree-based MST topology, this approach highlights an innovative perspective for enhancing SSM performance by rethinking the construction and definition of hidden states.


\subsection{Scan Strategy}\label{sec:scanstrategy}

\begin{table}[]
\renewcommand{\arraystretch}{1.33} 
\caption{This table summarizes a total of 44 different scan strategies utilized in the field of remote sensing. All acronyms are clearly illustrated in Fig. \ref{fig:scan_1_pre}-\ref{fig:multimodality}. The terms WB and BB denote 'Within Block' and 'Between Block', respectively, and are discussed in detail in Section \ref{sec:BB}. Arrows in the direction column indicate the starting token and initial scan directions, which are further explained in Section \ref{sec:scan directions}.}
\resizebox{\linewidth}{!}{ 
\begin{tabular}{cccccc|l}
\hline
\multicolumn{6}{c|}{Scan Strategy} & \multirow{2}{*}{Related Work}     \\ \cline{1-6}
\multicolumn{1}{c|}{Bran}    & Pre  & Sampling     & Direction       & Pattern  & Post   &\\ \hline
\multicolumn{7}{l}{\cellcolor[HTML]{C0C0C0}{\color[HTML]{000000} \textbf{Vallian Scan-like Strategies}}}   \\ \hline
\multicolumn{1}{c|}{-} & -  & - & $\nwarrow (\rightarrow)$ & Z  & - &  
    \begin{tabular}[c]{@{}l@{}} \cite{8-IMDCD}\cite{27-HLMamba}\cite{62-Mamba-MOC} \cite{5-S2CrossMamba} \\
    \cite{6-CDMamba,13-Samba,33-SpectralMamba,39-MSTFNet,42-PyramidMamba,43-MorpMamba,45-DualMamba,49-SS-Mamba,52-SSRFN,53-ConvMambaSR,61-WaveMamba,63-GraphMamba,74-SITSMamba,75-ES-HS-FPN,78-CSMN,89-HSIRMamba,101-MambaHSI,103-SCMamba }
    \end{tabular}
    \\ \hline
\multicolumn{1}{c|}{-} & -  & Window & $\nwarrow (\rightarrow)$ & Z & - & \cite{82-HDMba}                       \\ \hline
\multicolumn{1}{c|}{-} & - & -       & $\swarrow (\downarrow)$  & Hibert & PC &  \cite{95-LDMNet}   \\ \hline
\multicolumn{1}{c|}{-} & WB: CIAM  & - & $\nwarrow (\rightarrow)$ & Z & - & \cite{15-CMS2I-Mamba} \\ \hline
\multicolumn{7}{l}{\cellcolor[HTML]{C0C0C0}{\color[HTML]{000000} \textbf{Bidirectional Scan-like Strategies}}}   \\ \hline
\multicolumn{1}{c|}{-}  & - & -  & $\nwarrow\searrow (\rightarrow)$ & Z  & - &  \cite{28-FusionMamba,31-HyperMamba,47-SegMamba-OS,48-RSCaMa,64-S2Mamba, 67-DTAM,76-MamTrans,85-YOLO-Mamba,93-HSIMamba, 113-TrackingMamba} \\  \hline 
\multicolumn{1}{c|}{-}  & - & -  & $\nwarrow\searrow (\rightarrow)$ & Z  & PC &  \cite{107-MLMamba }                 \\ \hline 
\multicolumn{1}{c|}{-} & \multicolumn{1}{l}{WB: SaT-1} & -  & $\leftarrow\rightarrow (\rightarrow)$ & Z & - & \cite{29-MSFMamba,45-DualMamba,46-LE-Mamba, 64-S2Mamba } \\ \hline
\multicolumn{1}{c|}{-} & \multicolumn{1}{l}{WB: SaT-2} & -   & $\leftarrow\rightarrow (\rightarrow)$& Z &- & \cite{38-MambaLG} \\ \hline
\multicolumn{1}{c|}{-} & -   & -  & $\nwarrow (\rightarrow\downarrow)$ & Z    & - & \cite{35-AFA-Mamba}  \\ \hline
\multicolumn{1}{c|}{-} & - & - & $\nwarrow (\rightarrow\downarrow)$ & Local &  - & \cite{26-CVMH-UNet}\\ \hline
\multicolumn{1}{c|}{-} & \multicolumn{1}{l}{WB: GSRM} & - & $\nwarrow\searrow (\rightarrow)$ & Z & - & \cite{111-HSRMamba} \\ \hline
\multicolumn{1}{c|}{-}  & - & -  & $\nwarrow\searrow (\rightarrow)$ & Z  & ConN &  \cite{12-DMM, 14-ChangeMamba, 113-TrackingMamba}    \\ \hline 
\multicolumn{7}{l}{\cellcolor[HTML]{C0C0C0}{\color[HTML]{000000} \textbf{SS2D Scan-like Strategies}}}   \\ \hline
\multicolumn{1}{c|}{-} & -      & -      & $\nwarrow\searrow (\rightarrow\downarrow)$      & Z  & - & 
    \begin{tabular}[c]{@{}l@{}@{}@{}}
    \cite{12-DMM, 3-RTMamba, 2-RS3Mamba, 32-RemoteDet-Mamba, 37-HSDet-Mamba} 
    \\
    \cite{4-FreMamba, 7-RSDehamba, 28-FusionMamba, 17-RSMamba, 19-Mamba-CR}, \cite{84-COMO}, \cite{47-SegMamba-OS}
    \\ 
    \cite{112-RSVMamba}, \cite{49-SS-Mamba},  \cite{53-ConvMambaSR}, \cite{75-ES-HS-FPN}
    \\
    \cite{31-HyperMamba, 15-CMS2I-Mamba}
    \cite{14-ChangeMamba, 16-DC-Mamba, 18-UNetMamba, 22-Mamba-Diffusion,23-MFMamba, 25-CM-UNet, 30-Res-Mamba, 34-MambaFormer, 80-FMambaIR,83-MTIE-Net,86-Prompt-Mamba,87-Mamba-UAV-SegNet,91-SDMSPan,97-MGMF,99-MiM-ISTD,102-LCCDMamba,104-HMCNet,105-Mamba-MDRNet,108-SatMamba, 40-PPMamba,44-UVMSR,60-SSUM,69-EGCM-UNet,72-RFCC,41-MF-VMamba} \end{tabular} \\ \hline
\multicolumn{1}{c|}{-} & -      & ICG    & $\nwarrow\searrow (\rightarrow\downarrow)$  & Z  &  - & \cite{53-ConvMambaSR} \\ \hline
\multicolumn{1}{c|}{-} & -      & -      & $\nwarrow\searrow (\rightarrow\downarrow)$       & S  &  - & \cite{85-YOLO-Mamba,96-MaDiNet} \\ \hline
\multicolumn{1}{c|}{-} & -      & -      & $\nwarrow\searrow (\rightarrow\downarrow)$  & Local   & - & \cite{46-LE-Mamba}   \\ \hline
\multicolumn{1}{c|}{-}  & \multicolumn{1}{l}{WB: LP} & -  & $\nwarrow\searrow (\rightarrow\downarrow)$ & Z & - & \cite{81-ColorMamba}\\ \hline
\multicolumn{1}{c|}{-}  & \multicolumn{1}{l}{WB: SE}  & - & $\nwarrow\searrow (\rightarrow\downarrow)$ & Z &  - & \cite{88-VMambaSCI}            \\ \hline
\multicolumn{1}{c|}{-} & -      & -      & $\nwarrow\searrow (\rightarrow\downarrow)$      & Z  & ConN & \cite{14-ChangeMamba} \\ \hline
\multicolumn{1}{c|}{-} & -      & -      & $\nwarrow\searrow (\rightarrow\downarrow)$      & Z  & ConC & \cite{14-ChangeMamba, 102-LCCDMamba} \\ \hline
\multicolumn{1}{c|}{-} & -      & -      & $\nwarrow\searrow (\rightarrow\downarrow)$      & Z  & CConN & \cite{14-ChangeMamba, 110-CD-Lamba, 102-LCCDMamba} \\ \hline
\multicolumn{1}{c|}{-} & -      & -      & $\nwarrow\searrow (\rightarrow\downarrow)$      & Z  & E1DS & \cite{22-Mamba-Diffusion} \\ \hline

\multicolumn{7}{l}{\cellcolor[HTML]{C0C0C0}{\color[HTML]{000000} \textbf{Novel Scanning Strategies}}}   \\ \hline

\multicolumn{1}{c|}{-} & -  & -  & \begin{tabular}[c]{@{}c@{}}Center Token\\ Clockwise\end{tabular} & Helical  &  - & \cite{59-STMamba}  \\ \hline
\multicolumn{1}{c|}{-} & -  & -  & $\nwarrow (\rightarrow)$ & Tree &  - &  \cite{68-TTMGNet}                 \\ \hline
\multicolumn{1}{c|}{-} & -  & -  & 12 Directions & S & - & \cite{79-SSUMamba}  \\ \hline
\multicolumn{1}{c|}{\multirow{2}{*}{Para}} & \multirow{2}{*}{-} & \multirow{2}{*}{-} & \multirow{2}{*}{$\nwarrow\searrow (\rightarrow\downarrow)$} & Z  & \multirow{2}{*}{}    \\
\multicolumn{1}{c|}{}   &  & &   & Diagonal   &   - &     \cite{1-RS-Mamba,21-UV-Mamba,56-PPMamba }                 \\ \hline

\multicolumn{7}{l}{\cellcolor[HTML]{C0C0C0}{\color[HTML]{000000} \textbf{Two Paradigms of Omnidirectional Scan Strategy}}}   \\ \hline

\multicolumn{1}{c|}{-} & -  & -  & $\nwarrow\nearrow\searrow\swarrow (\rightarrow\downarrow)$  & S  & - & \cite{36-MiM}                       \\ \hline
\multicolumn{1}{c|}{\multirow{2}{*}{Para}} & \multirow{2}{*}{-} & \multirow{2}{*}{-} & $\nwarrow\searrow (\rightarrow)$ & Z & \multirow{2}{*}{}       \\
\multicolumn{1}{c|}{}  &         &   & $\nwarrow (\rightarrow)$  & Random    &  - & \cite{9-RSMamba,15-CMS2I-Mamba,109-UrbanSSF }                      \\ \hline

\multicolumn{7}{l}{\cellcolor[HTML]{C0C0C0}{\color[HTML]{000000} \textbf{Random Scan Strategy + Other Scan Strategies}}}   \\ \hline

\multicolumn{1}{c|}{\multirow{2}{*}{Para}} & -  & -  & $\nwarrow\searrow (\rightarrow)$ & Z  & \\
\multicolumn{1}{c|}{}  & \multicolumn{1}{l}{WB: SaT-1}  & -   & $\leftarrow (\rightarrow)$ & Random & - &  \cite{15-CMS2I-Mamba} \\ \hline
\multicolumn{1}{c|}{\multirow{2}{*}{Para}} & \multirow{2}{*}{-}  & \multirow{2}{*}{-} & $\nwarrow\searrow (\rightarrow\downarrow)$ & Z &    \\
\multicolumn{1}{c|}{}                      &                     &                    & $\nwarrow (\rightarrow)$               & Random& - & \cite{92-MambaDS}  \\ \hline

\multicolumn{7}{l}{\cellcolor[HTML]{C0C0C0}{\color[HTML]{000000} \textbf{Multi-Router(Sampling/Directions/Patterns) Scanning Strategies}}}   \\ \hline

\multicolumn{1}{c|}{\multirow{2}{*}{Para}} & \multirow{2}{*}{-}  & \multirow{2}{*}{-} & \multirow{2}{*}{$\nwarrow (\rightarrow\downarrow)$} & Z & \multirow{2}{*}{}       \\
\multicolumn{1}{c|}{}                      &                     &                   &         & S         &  - & \cite{24-LightMamba}   \\ \hline
\multicolumn{1}{c|}{\multirow{3}{*}{Para}} & \multirow{3}{*}{-} & \multirow{3}{*}{-}      & \multirow{3}{*}{$\nwarrow (\rightarrow)$} & Z    & \multirow{3}{*}{}       \\
\multicolumn{1}{c|}{}                    &                    &                         &                                        & S    &                         \\
\multicolumn{1}{c|}{}                    &                    &                         &       & Diagonal  & - & \cite{106-Weaba}       \\ \hline
\multicolumn{1}{c|}{\multirow{2}{*}{Para}} & \multirow{2}{*}{-} & \multirow{2}{*}{-}& \multirow{2}{*}{$\nwarrow\searrow (\rightarrow)$} & Z  & \multirow{2}{*}{}       \\
\multicolumn{1}{c|}{}                      &                    &                   &    & Diagonal                     &  - &  \cite{26-CVMH-UNet}\\ \hline
\multicolumn{1}{c|}{\multirow{2}{*}{Para}} & \multirow{2}{*}{-} & \multirow{2}{*}{Atrous} & $\nwarrow (\rightarrow)$ & Z & \multirow{2}{*}{}       \\
\multicolumn{1}{c|}{}                      &                    &                         & $\nwarrow (\downarrow)$  & Z & - &  \cite{94-HRMamba-YOLO}\\ \hline

\multicolumn{7}{l}{\cellcolor[HTML]{C0C0C0}{\color[HTML]{000000} \textbf{Sequential Multi-Router Scanning Strategies}}}   \\ \hline

\multicolumn{1}{c|}{\multirow{2}{*}{Seq}}  & \multirow{2}{*}{-} & \multirow{2}{*}{-} & $\nwarrow (\rightarrow)$ & Z  & \multirow{2}{*}{}       \\
\multicolumn{1}{c|}{}                      &                    &                    & $\searrow (\rightarrow)$ & Z  &  - &  \cite{70-TransMamba} \\ \hline
\multicolumn{1}{c|}{\multirow{2}{*}{Seq}}  & \multirow{2}{*}{-} & \multirow{2}{*}{-} & $\nwarrow\swarrow (\rightarrow\downarrow)$ & Z  & \multirow{2}{*}{}       \\
\multicolumn{1}{c|}{}   &  &   & $\nearrow\searrow (\rightarrow\downarrow)$ & S  & - & \cite{58-HSIDMamba}   \\ \hline
\multicolumn{1}{c|}{\multirow{2}{*}{Seq}}  & \multirow{2}{*}{-} & Window & $\nwarrow (\rightarrow)$ & S  &                         \\
\multicolumn{1}{c|}{}     &       & -      & $\nwarrow (\rightarrow)$ & S  & - & \cite{38-MambaLG}   \\ \hline

\multicolumn{7}{l}{\cellcolor[HTML]{C0C0C0}{\color[HTML]{000000} \textbf{Novel Preprocessing Strategies}}}   \\ \hline

\multicolumn{1}{c|}{\multirow{2}{*}{Para}}  & -                  & -   & $\nwarrow\searrow (\rightarrow)$  & S   &                         \\
\multicolumn{1}{c|}{}        & \multicolumn{1}{l}{WB: SaT-1}    & -   & $\leftarrow \rightarrow (\rightarrow)$  & S   & - & \cite{55-3DSS-Mamba} \\ \hline
\multicolumn{1}{c|}{\multirow{2}{*}{Para}}  & \multicolumn{1}{l}{WB: $\text{MSD}^{1}$} & - & $\nwarrow (\rightarrow\downarrow)$ & S  &   \\
\multicolumn{1}{c|}{}   & \multicolumn{1}{l}{WB: $\text{MSD}^{2}$} & - & $\searrow (\rightarrow\downarrow)$ & S  & - & \cite{29-MSFMamba} \\ \hline
\multicolumn{1}{c|}{-}  & \multicolumn{1}{l}{WB: $\text{MSD}^4$} & -  & $\nwarrow (\rightarrow)$ & S   & - &  \cite{73-HTD-Mamba}  \\ \hline 
\multicolumn{1}{c|}{-}  & \multicolumn{1}{l}{WB: DTA} & -  & $\nwarrow\searrow(\rightarrow)$ & Z   &  - & \cite{67-DTAM }  \\ \hline 
\multicolumn{1}{c|}{\multirow{2}{*}{Seq}}  & \multicolumn{1}{l}{WB: CG}  & -   & $\nwarrow (\rightarrow)$  & S   &                         \\
\multicolumn{1}{c|}{}          & \multicolumn{1}{l}{WB: SaT-1+CG}        & -   & $\leftarrow (\rightarrow)$  & S   &  - & \cite{51-IGroupSS-Mamba} \\ \hline
\multicolumn{1}{c|}{-} & \multicolumn{1}{l}{BB: SsM}             
    & -   & $\nwarrow\searrow (\rightarrow\downarrow)$ & \begin{tabular}[c]{@{}c@{}}Nested \\ S-Shape\end{tabular} & - & \cite{90-MaIR} \\ \hline
\multicolumn{1}{c|}{-} & \multicolumn{1}{l}{BB: WSM} & -  & $\nwarrow (\rightarrow)$ & AR-Shape  & - & \cite{110-CD-Lamba} \\ \hline
\end{tabular}
}
\label{table:scanning}
\end{table}

Scan strategies have become a critical area of focus for enhancing Mamba-based models. A fundamental challenge in CV lies in effectively transforming 2D image features into the 1D sequences required by Mamba models. Several recent surveys \cite{survey1, survey3, survey4, survey5, surveyon2, survey6} have reviewed existing scanning strategies. Nevertheless, these surveys have two significant limitations: (1) Being published earlier, they do not comprehensively cover all current scanning strategies. (2) Most existing surveys simply enumerate scanning techniques without providing a systematic classification framework. Although two surveys \cite{survey5, survey6} made initial attempts to categorize scanning strategies, their classification methods have become inadequate due to the rapid and continuous evolution of scanning techniques.

\begin{figure*}[htbp]
\centering
\includegraphics[width=1.0\textwidth]{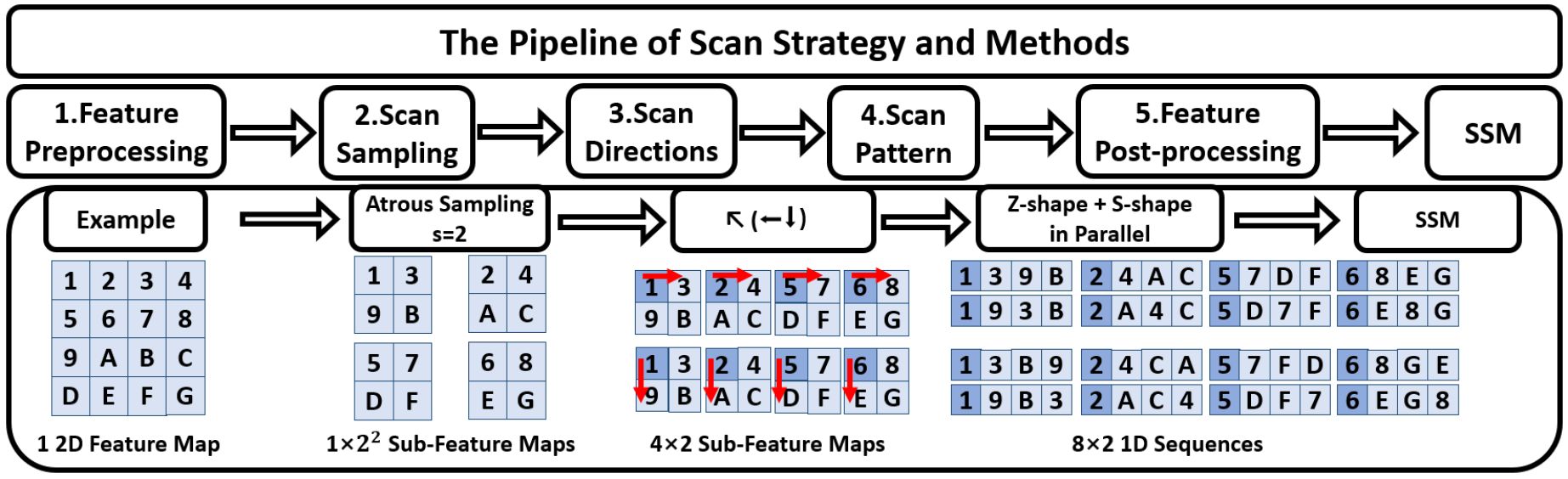}
\caption{Illustration of the scan strategy pipeline, comprising five key components, including feature preprocessing, scan sampling, scan directions, scan pattern, and feature post-processing. These five stages essentially transform a 2D feature map into multiple 1D sequences that conform to Mamba's processing architecture.
For clarity, an additional example is provided demonstrating the case without any preprocessing or postprocessing operations.}
\label{fig:scan_pipeline}
\end{figure*}

To overcome these limitations, this survey paper conducts a comprehensive review of all existing scanning strategies within the remote sensing domain and introduces an innovative, unified classification framework applicable to all current scanning methods. Specifically, we propose a detailed classification framework comprising five key components: {\it feature preprocessing methods}, {\it scan sampling methods}, {\it scan direction methods}, {\it scan pattern methods}, and {\it post-processing methods}. These components can be employed either sequentially or concurrently. 
Fig. \ref{fig:scan_pipeline} provides a visual representation of this scanning pipeline, summarizing existing methods within remote sensing and presenting an illustrative workflow. 
In total, 44 scanning strategies have been identified and systematically categorized in Tab. \ref{table:scanning}. It is important to note that all of these methods are designed for single-modality images. For multi-modality images, an additional feature preprocessing method and four feature post-processing methods are available, which are elaborated in Sections \ref{sec:mulplePre} and \ref{sec:mulplePost}, respectively, and visually represented in Fig. \ref{fig:multimodality}.

Here, we define image height as $H$, image width as $W$, and latent dimension (or spectral dimension for hyperspectral and multispectral data) as $C$. Upon converting a 2D feature map into a 1D sequence, the number of resulting tokens is calculated as $N = H \times W$. To emphasize the contribution of each method while preserving naming simplicity and consistency, we may modify the name of original name.

\subsubsection{\textbf{Feature Preprocessing}} \label{sec:preprocess}

\begin{figure*}[htbp]
\centering
\includegraphics[width=1.0\textwidth]{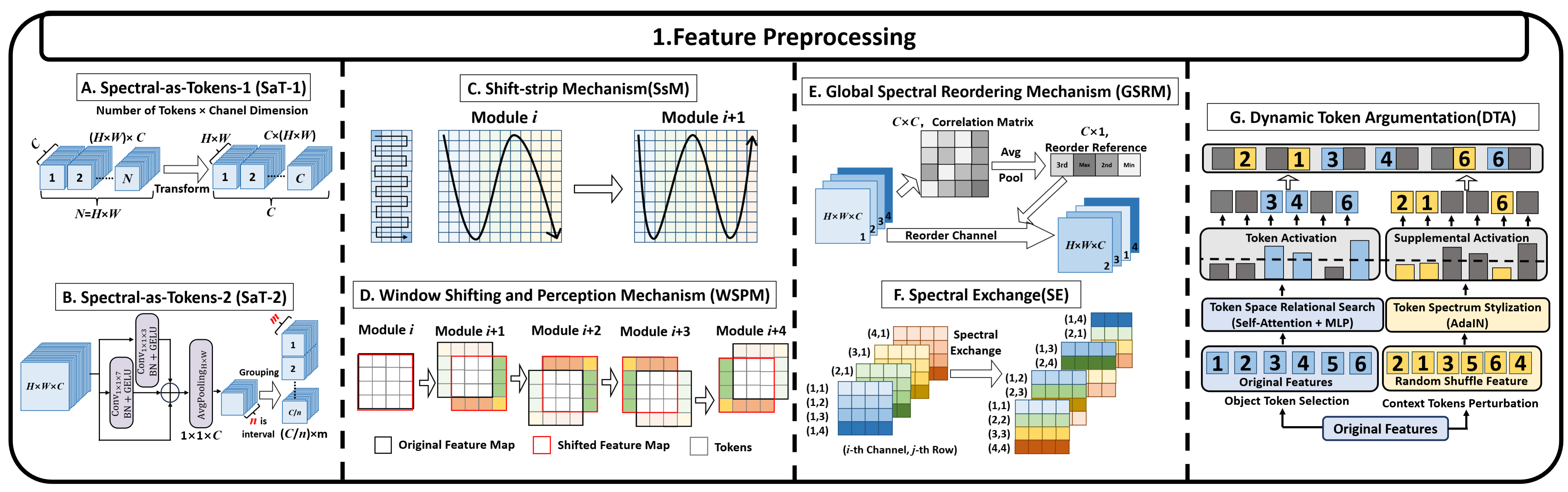}
\caption{The seven feature preprocessing methods for scan strategy. For clarity, two additional preprocessing methods are visually shown in Fig.~\ref{fig:scan_1_pre2}.}
\label{fig:scan_1_pre}
\end{figure*}

Feature preprocessing is distinct from conventional feature extraction methods such as CNNs and ViTs. Instead of extracting features directly from input data, feature preprocessing focuses on modifying existing feature maps through various operations, including transformations, spectral exchanges, padding, and others. These preprocessing techniques can lead to significant differences in the output 1D sequences, thus impacting the effectiveness and performance of the overall scan strategies. In total, nine distinct preprocessing methods have been identified and classified into four categories.

\paragraph{Spectral-as-Tokens (SaT)}
In remote sensing, hyperspectral and multispectral data provide abundant spectral information. Improving spectral modeling capabilities can substantially enhance the performance of deep learning models. This paper refers to these approaches as Spectral-as-Tokens (SaT). The SaT methods enable SSMs to model extensive spectral dependencies, converting 2D or 3D features naturally into 1D sequential formats. The first variant, termed \textbf{SaT-1} \cite{15-CMS2I-Mamba, 29-MSFMamba, 45-DualMamba, 46-LE-Mamba, 55-3DSS-Mamba, 51-IGroupSS-Mamba}, rearranges the spectral ($C$) and spatial ($H \times W$) dimensions to form a feature map comprising $C$ tokens, each possessing an $H \times W$ latent representation, as depicted in Fig. \ref{fig:scan_1_pre}.A. This permutation naturally facilitates SSMs in effectively modeling spectral correlations. The second variant, called \textbf{SaT-2} \cite{38-MambaLG}, employs multi-scale 3D NNs to generate feature representations. These features are subsequently averaged and partitioned into $n$ intervals, each having $m$ latent dimensions, as illustrated in Fig. \ref{fig:scan_1_pre}.B.

\paragraph{Feature Map Shift between Blocks} \label{sec:BB}
While SaT methods operate within blocks (WB), this category focuses on the feature transformation between blocks (BB), enabling Mamba models to attend to varying feature regions across multiple contiguous blocks.
MaIR \cite{90-MaIR} introduces the \textbf{Stripe-shift Mechanism (SsM)}, which alters the scanning region for its nested S-shape scan pattern to create diverse 1D sequences, as depicted in Fig.~\ref{fig:scan_1_pre}.C. Similarly, CDLamba \cite{110-CD-Lamba} proposes the \textbf{Window Shifting and Perception Mechanism (WSPM)}, shifting feature windows similar to Swin-Transformer \cite{SwinTransformer}, thereby generating diverse window configurations as shown in Fig. \ref{fig:scan_1_pre}.D. These techniques enhance the model's capacity to focus on distinct regions between blocks, significantly improving feature representation.

\paragraph{Feature Map Rearranging}
Unlike Transformers \cite{Transformer, DBLP:journals/pami/LiDYZPCCLL24} processing tokens in parallel, Mamba performs sequential computation, meaning that the initial scan order and the starting token influence subsequent token interactions. This category aims to optimize spatial-spectral feature extraction through feature map rearrangement \cite{111-HSRMamba, 88-VMambaSCI, 67-DTAM}.
HSRMamba \cite{111-HSRMamba} introduces the \textbf{Global Spectral Reordering Mechanism (GSRM)}, which rearranges spectral features based on the average correlation matrix values between spectrum, ensuring that highly correlated pixels are positioned closer together, as illustrated in Fig. \ref{fig:scan_1_pre}.E. This improves long-range spectral feature modeling.
VMambaSCI \cite{88-VMambaSCI} proposes \textbf{Spectral Exchange (SE)}, which reorders spectral features using a predefined sequence, as shown in Fig. \ref{fig:scan_1_pre}.F. This enhances the long-range modeling of spectral features within the same spatial channel.
DTAM \cite{67-DTAM} introduces \textbf{Dynamic Token Argumentation (DTA)} to enhance global feature learning. As depicted in Fig. \ref{fig:scan_1_pre}.G, DTA consists of two routes: the left route activates object-related tokens while masking the others, and the right route activates contextual tokens while using random shuffling and Adaptive Instance Normalization (AdaIN) \cite{AdaIN} for augmentation. This approach helps the model focus on meaningful object-related features while mitigating spectral variability caused by acquisition conditions.

\begin{figure}[htbp]
\centering
\includegraphics[width=0.48\textwidth]{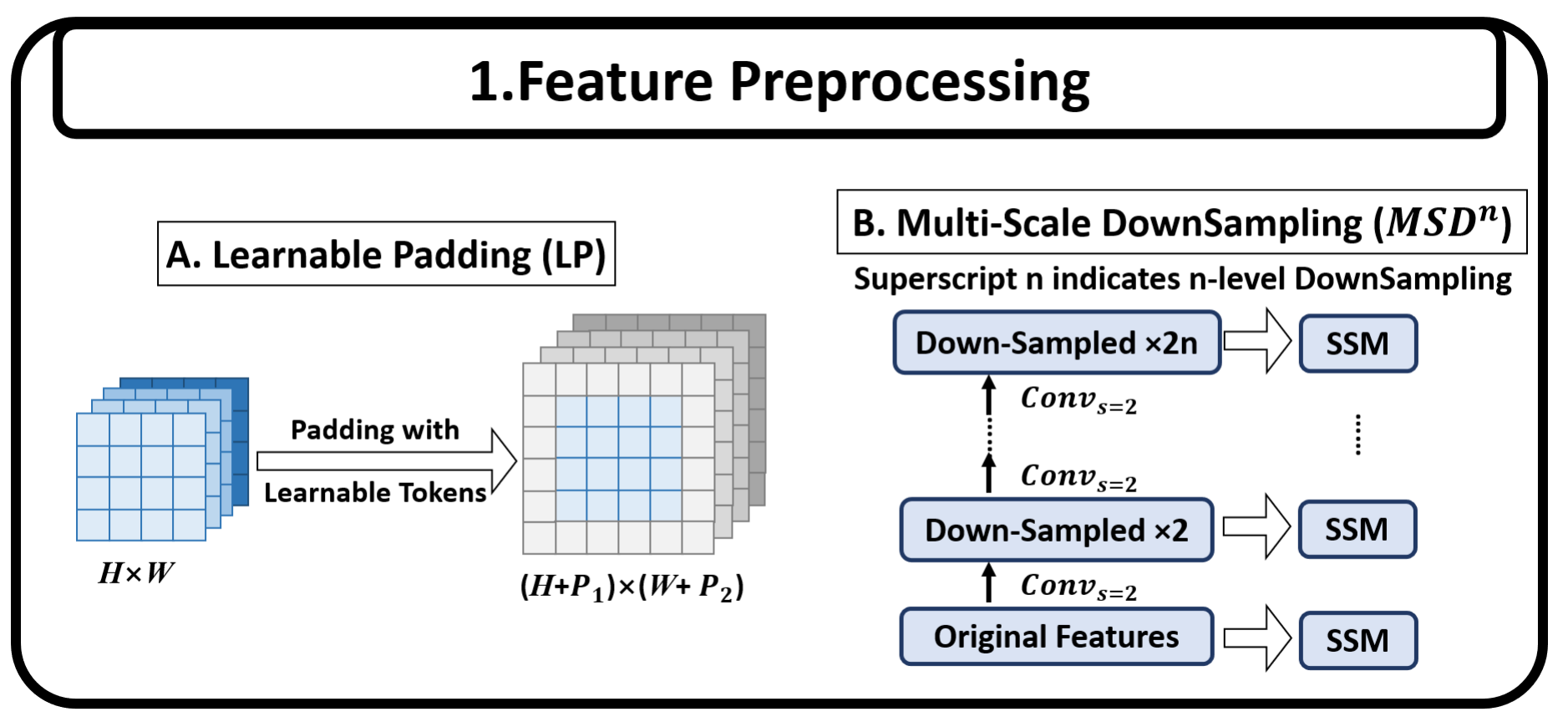}
\caption{Two additional feature preprocessing methods for scan strategy.}
\label{fig:scan_1_pre2}
\end{figure}

\paragraph{Other Preprocessing Operations}
Additional preprocessing techniques further aim to enhance Mamba's long-range modeling capabilities. ColorMamba \cite{81-ColorMamba} identifies that certain scan patterns (discussed in Section \ref{sec:scan pattern}) can disrupt spatial relationships. To address this, it introduces \textbf{Learnable Padding (LP)}, embedding learnable tokens around feature maps, illustrated in Fig. \ref{fig:scan_1_pre2}.A. These tokens enhance spatial representations and maintain sequential continuity. MSFMamba \cite{29-MSFMamba} and HTD-Mamba \cite{73-HTD-Mamba} adopt \textbf{Multi-Scale Downsampling (MSD)}, which utilizes strided CNNs for feature downsampling, thereby improving long-range modeling of multi-scale features, as depicted in Fig. \ref{fig:scan_1_pre2}.B.

\subsubsection{\textbf{Scan Sampling}}
\begin{figure}[htbp]
\centering
\includegraphics[width=0.48\textwidth]{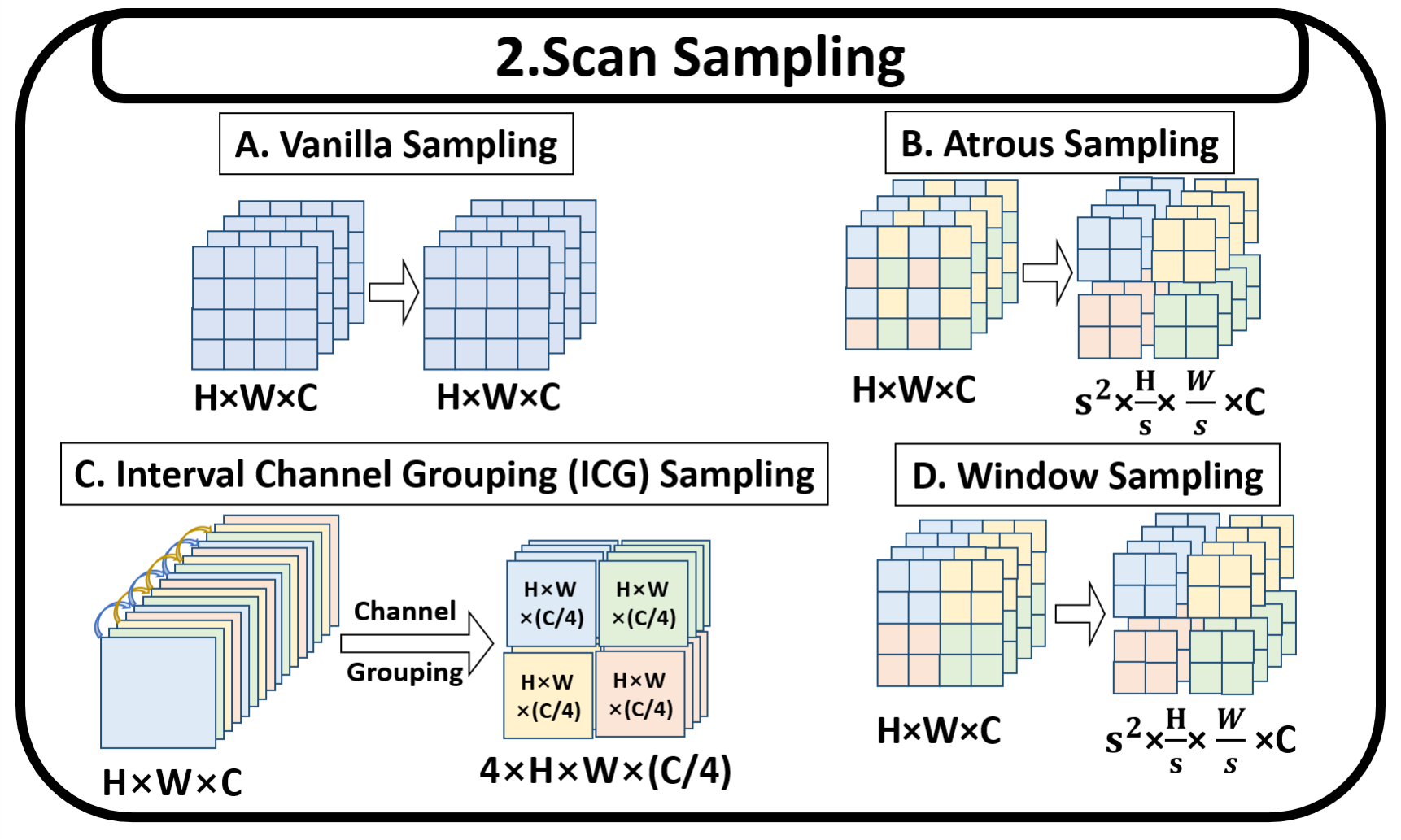}
\caption{The four scan sampling for the scan strategy.}
\label{fig:scan_2_sampling}
\end{figure}

Scan sampling refers to the process of sampling the original feature map into multiple non-overlapping sub-feature maps.
Most studies utilize \textbf{Vanilla Sampling}, which is essentially equivalent to not performing any sampling at all or represents a special case of other scan sampling methods, as illustrated in Fig. \ref{fig:scan_2_sampling}.A.
\textbf{Atrous Sampling} \cite{94-HRMamba-YOLO}, inspired by atrous CNNs \cite{atrousCNN}, samples feature maps at an interval of atrous step $s$, resulting in $s^2$ sub-feature maps with dimensions of $\frac{H}{s} \times \frac{W}{s} \times C$. This method effectively enhances the receptive field, as demonstrated in Fig. \ref{fig:scan_2_sampling}.B.
\textbf{Interval Channel Grouping Sampling (ICG)} \cite{51-IGroupSS-Mamba} selects feature maps at an interval of $g$ along the $C$ dimension, generating $g$ sub-feature maps, each with dimensions $H \times W \times \frac{C}{g}$. This method facilitates the construction of non-redundant global information within each group while reducing data dimensionality, thereby enhancing computational efficiency \cite{51-IGroupSS-Mamba}, as illustrated in Fig. \ref{fig:scan_2_sampling}.C.
\textbf{Window Sampling} \cite{38-MambaLG, 82-HDMba}, similar to the windowing mechanism in Swin-Transformer \cite{SwinTransformer}, partitions feature maps into several windows, each consisting of $n \times m$ tokens. Consequently, it produces $\frac{H}{n} \times \frac{W}{m}$ windows (sub-feature maps) with dimensions of $n \times m \times C$. This approach primarily focuses on modeling local features within individual windows, as depicted in Fig. \ref{fig:scan_2_sampling}.D.

\subsubsection{\textbf{Scan Directions}} \label{sec:scan directions}
Scan directions determine the starting token (typically a corner token or the center token) and the initial scanning trajectory (e.g., vertical, horizontal, clockwise, or counterclockwise) across all sub-feature maps. Since transforming 2D feature maps into 1D sequences inevitably disrupts spatial relationships, employing multiple scan directions helps improve the sequence model’s spatial understanding \cite{VisionMamba, VMamba}.

\begin{figure}[htbp]
\centering
\includegraphics[width=0.48\textwidth]{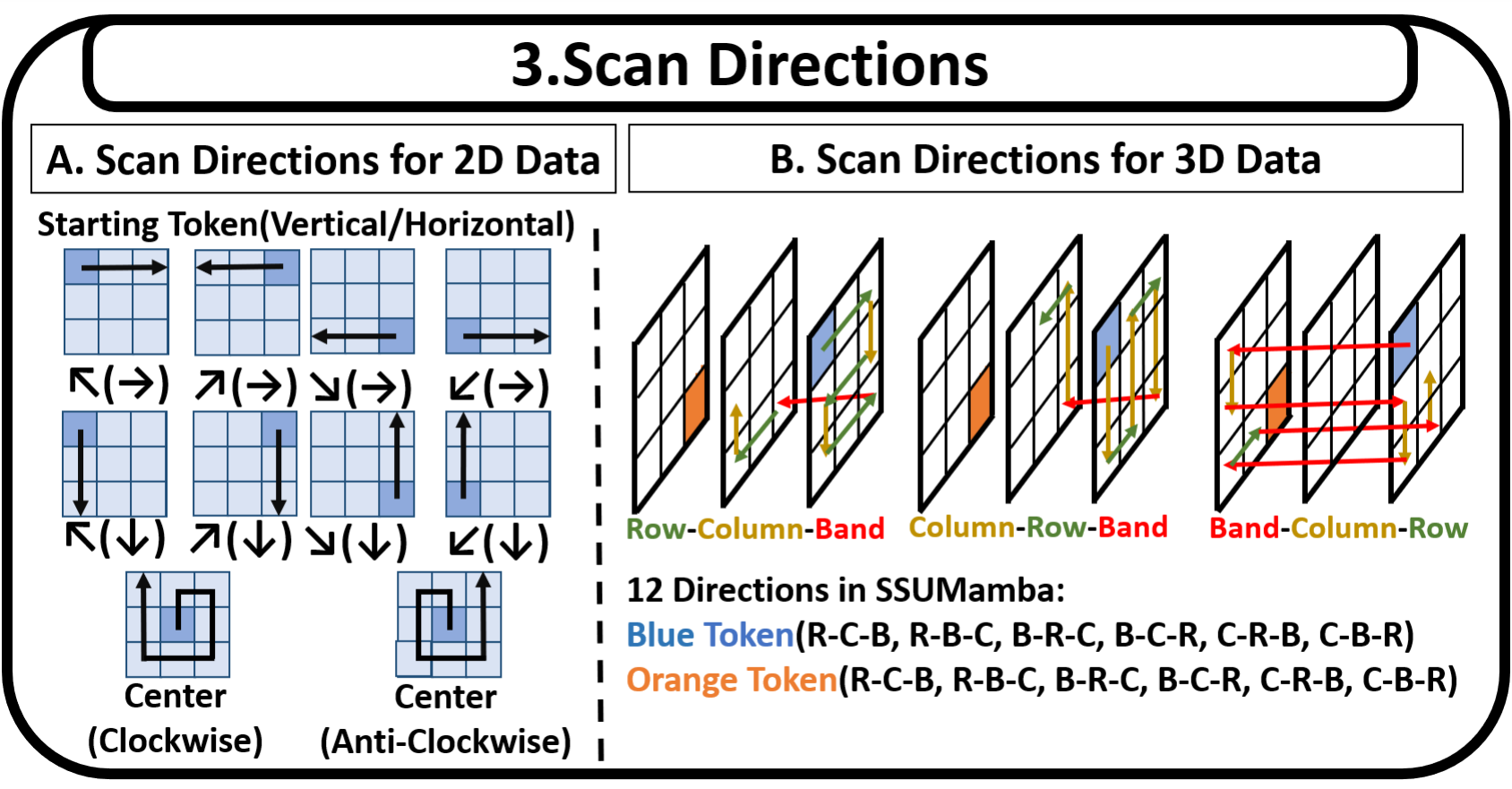}
\caption{The 11 scam directions for scan strategy.}
\label{fig:scan_3_directions}
\end{figure}

A scanning strategy can utilize $d$ different scan directions, resulting in $d$ times the number of original sub-feature maps. While scan sampling segments a feature map into multiple sub-feature maps without additional computational cost, increasing scan directions effectively replicates each sub-feature map $d$ times, leading to $O(d \times N)$ computational complexity.

As illustrated in Fig. \ref{fig:scan_3_directions}.A and summarized in Table \ref{table:scanning}, diagonal arrows indicate the starting corner token, such as $\nwarrow$ representing the top-left corner. Horizontal ($\rightarrow$) and vertical ($\downarrow$) arrows denote the initial scanning directions. For instance, Vision Mamba (Vim) \cite{VisionMamba} introduces a bi-directional scan, starting from the top-left and bottom-right tokens vertically, which is recorded as $\nwarrow\searrow (\rightarrow)$. Bi-directional scan results in twice of the original feature maps. 
Similarly, VMamba \cite{VMamba} proposes the VSS method, initiating scanning from both the top-left and bottom-right tokens in both vertical and horizontal directions, recorded as $\nwarrow\searrow (\rightarrow \downarrow)$. VSS method results in four times the original feature maps. A special case involves scanning from the center token in a clockwise direction \cite{59-STMamba}, specifically designed to align with a designated helical scan pattern (discussed in Section \ref{sec:scan pattern}). Our analysis of 2D feature maps reveals a total of 11 scanning directions.

This concept can be extended to 3D data, as illustrated in Fig. \ref{fig:scan_3_directions}.B. In 3D feature maps, eight regular corner tokens are available as initial points for scanning, with three primary scanning directions: vertical (along the $H$ dimension), horizontal (along the $W$ dimension), and spectral (along the spectral dimension), yielding $3 \times 8$ fundamental scanning paths. Incorporating additional starting points, such as the central token, and further directions like clockwise and counterclockwise scanning significantly expands the potential number of 3D scan directions. Within remote sensing contexts, SSUMamba \cite{79-SSUMamba} is currently the sole Mamba-based approach treating hyperspectral data in 3D, introducing the first 3D data scan strategy method, which employs 12 of these scanning directions, as illustrated in Fig. \ref{fig:scan_3_directions}.B.

Similarly, certain preprocessing methods, such as SaT, can directly transform 2D feature maps into 1D sequences, bypassing some subsequent scan steps. For 1D sequences, scanning options include two starting tokens, namely the first token ($\leftarrow$) and the last token ($\rightarrow$). In addition, the initial direction is from left to right if the starting token is the first, and is from right to left if the starting token is the last, which is recorded as $\rightarrow$.

\subsubsection{\textbf{Scan Patterns}} \label{sec:scan pattern}
\begin{figure}[htbp]
\centering
\includegraphics[width=0.48\textwidth]{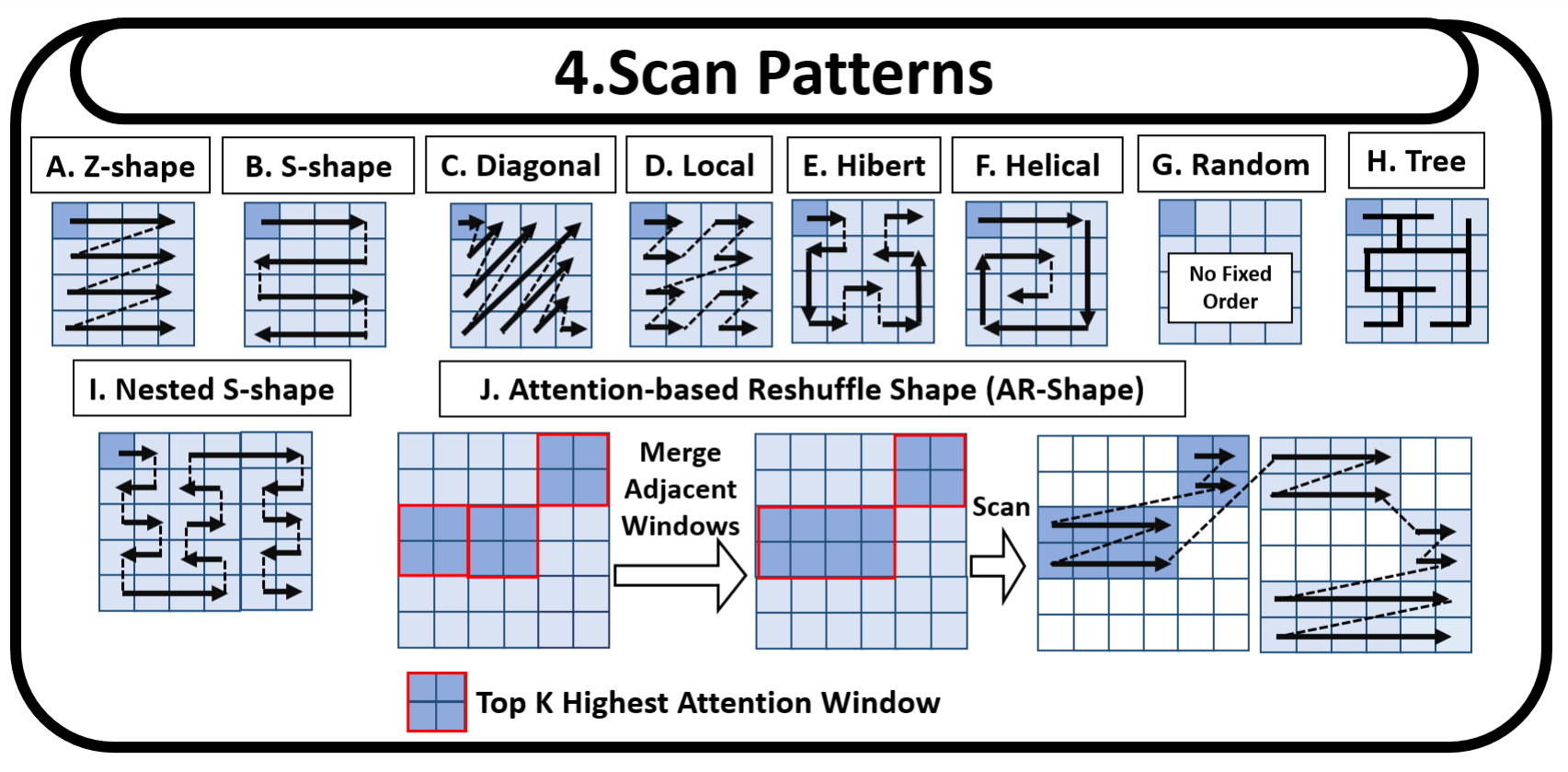}
\caption{The ten scan patterns for the scan strategy.}
\label{fig:scan_4_pattern}
\end{figure}

After completing the previous steps, the subsequent step involves employing suitable scan patterns to systematically convert the 2D feature map into 1D sequences. Various scan patterns provide unique benefits tailored to specific applications. In remote sensing, 10 primary scan patterns are commonly used, which are visually illustrated in Fig. \ref{fig:scan_4_pattern}.

\paragraph{Vanilla Scan Pattern (Z-Shape)}
The Z-shape pattern \cite{VisionMamba, VMamba} is the most straightforward transformation, achieved by directly flattening the feature map into a 1D sequence, as shown in Fig. \ref{fig:scan_4_pattern}.A. Although straightforward and commonly employed, this pattern compromises spatial relationships in the original 2D feature map. In particular, transitions between the last token of each row and the first token of the subsequent row lack spatial continuity, possibly resulting in diminished semantic context.

\paragraph{Continuous Pattern (S-Shape)}
Introduced by PlainMamba \cite{plainmamba}, the S-shape pattern, depicted in Fig. \ref{fig:scan_4_pattern}.B, preserves spatial continuity by ensuring that scanning progresses smoothly across rows. This effectively resolves the spatial discontinuity issue characteristic of the Z-shape pattern, making it especially advantageous for high-resolution remote sensing data, where preserving spatial consistency significantly enhances performance.

\paragraph{Diagonal Pattern}
The diagonal scanning pattern, first adopted by VMambaIR \cite{VMambaIR}, enhances spatial connectivity by following a diagonal trajectory, as illustrated in Fig. \ref{fig:scan_4_pattern}.C. This approach is particularly effective for capturing long-range dependencies in high-resolution remote sensing imagery.

\paragraph{Local Pattern}
Local Mamba \cite{LocalMamba} employs a local pattern, first processing tokens within individual windows before moving to adjacent windows, as shown in Fig. \ref{fig:scan_4_pattern}.D. This method is advantageous for capturing fine-grained local features. However, a drawback is that it disrupts the continuity between different windows, which may limit its ability to model global structures.

\paragraph{Hilbert Pattern}
Hilbert scanning \cite{95-LDMNet} utilizes the Hilbert curve, a space-filling trajectory that recursively subdivides the image while maintaining spatial coherence, as illustrated in Fig. \ref{fig:scan_4_pattern}.E. This pattern effectively preserves local adjacency relationships, enhancing the spatial consistency of the generated 1D sequence.

\paragraph{Helical Pattern}
The helical pattern \cite{59-STMamba} initiates scanning from the center token and proceeds outward in a clockwise manner, as depicted in Fig. \ref{fig:scan_4_pattern}.F. This method is particularly useful in remote sensing tasks where classification is based on the central pixel of a high-resolution patch. By prioritizing the central region, the helical pattern ensures more precise feature representation.

\paragraph{Random Pattern}
The random pattern \cite{9-RSMamba, 15-CMS2I-Mamba, 109-UrbanSSF, 92-MambaDS} disrupts sequential order by randomly rearranging tokens to enhance positional invariance, as shown in Fig. \ref{fig:scan_4_pattern}.G. This approach is often used in conjunction with other scan patterns to improve robustness against spatial transformations.

\paragraph{Tree Pattern}
The tree-based pattern \cite{68-TTMGNet} constructs a minimum spanning tree (MST) by eliminating edges with low cosine similarity, as demonstrated in Fig. \ref{fig:scan_4_pattern}.H. The 1D sequence is then generated through a Breadth-First Search (BFS) traversal of the tree. This method enhances global information extraction and generalization across different input structures.

\paragraph{Nested S-Shape Pattern}
The nested S-shape pattern, proposed by MaIR \cite{90-MaIR}, integrates the principles of the S-shape and local patterns. It partitions the feature map into non-overlapping stripes and applies S-shape scanning within each stripe, as depicted in Fig. \ref{fig:scan_4_pattern}.I. Additionally, the shift-strip mechanism (discussed in Section \ref{sec:BB}) further refines this approach to enhance feature extraction.

\paragraph{Attention-based Reshuffling Pattern (AR-Shape)}
The AR-Shape pattern, introduced by CD-Lamba \cite{110-CD-Lamba}, dynamically prioritizes scan order based on an attention mechanism, as shown in Fig. \ref{fig:scan_4_pattern}.J. A 4$\times$4 average pooling operation generates attention scores, which determine the order in which windows are merged and scanned. This approach preserves spatial continuity in semantically critical regions, such as roads or buildings within remote sensing imagery, thereby making it especially suitable for change detection (CD) applications~\cite{DBLP:journals/remotesensing/ChengHLLXZZX24}. 

\subsubsection{\textbf{Feature Post-Processing}} \label{sec:post}
Feature post-processing involves operations conducted after acquiring the 1D sequence. In some cases, post-processing can function similarly to pre-processing, depending on the chosen feature preparation methods, scan sampling approaches, scan directions, and scanning patterns. Nevertheless, it is crucial to note that further operations may still be applied after generating the 1D sequences to enrich the diversity of scan strategies. For single-modality images, only one post-processing technique, Pyramid Concatenation (PC), introduced by MLMamba \cite{107-MLMamba} and LDMNet \cite{95-LDMNet}, which concatenates multi-scale feature representations before feeding them into SSMs. Compared with MSD methods, PC inherently facilitates one SSM block in capturing long-range dependencies across features of various scales, thus improving the overall effectiveness of the model. Methods tailored for multimodal images will be detailed in Section \ref{sec:mulplePost}.

\begin{figure}[htbp]
\centering
\includegraphics[width=0.48\textwidth]{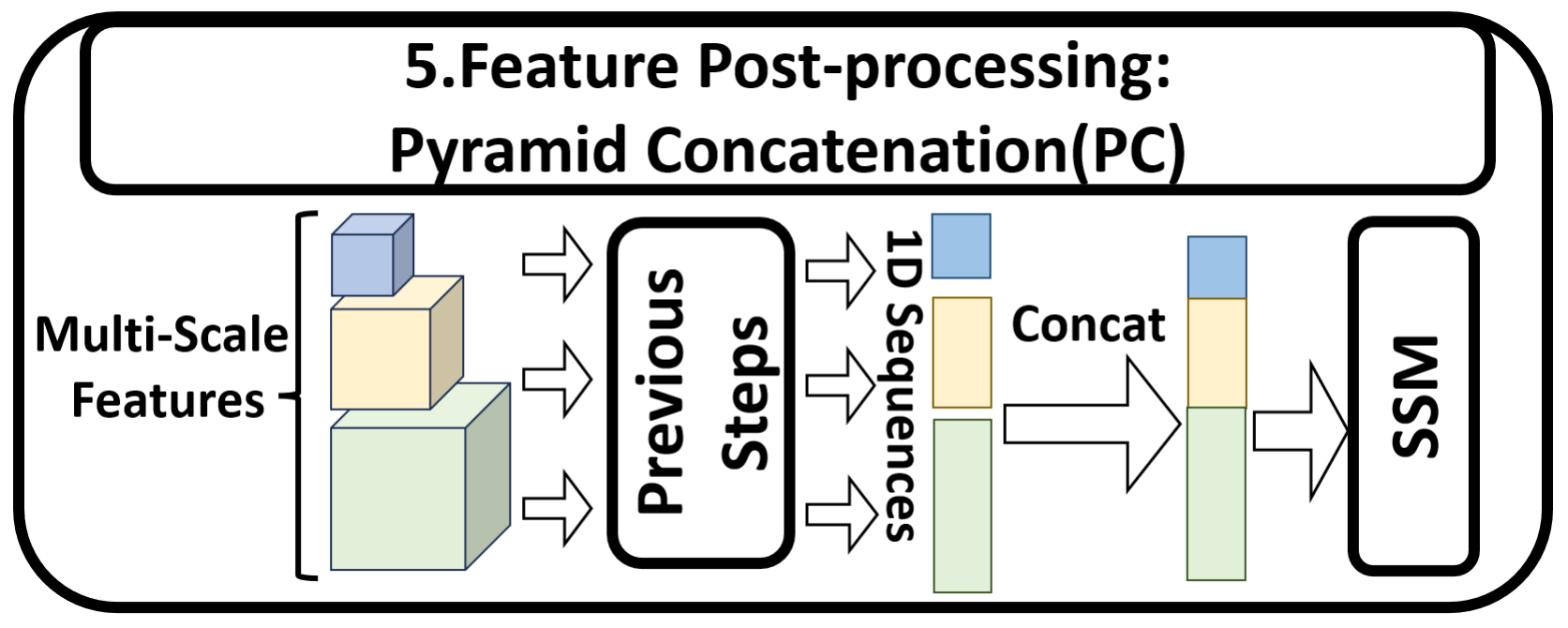}
\caption{One feature post-processing method for scan strategy.}
\label{fig:scan_5_post}
\end{figure}



\subsection{Multi-Modal and Bi-Temporal Feature Interaction via Mamba} \label{sec:multimodality}
Numerous studies \cite{5-S2CrossMamba, 12-DMM, 14-ChangeMamba, 15-CMS2I-Mamba, 22-Mamba-Diffusion, 28-FusionMamba, 29-MSFMamba, 32-RemoteDet-Mamba, 47-SegMamba-OS, 48-RSCaMa, 84-COMO,110-CD-Lamba, 113-TrackingMamba} have leveraged multimodal and bi-temporal data, such as conventional RGB images, LiDAR point clouds and infrared images, to enhance performance. The bi-temporal data refers to two images captured at different time in the same location, normally for the Change Detection (CD) task.

A crucial aspect of multimodal and bi-temporal learning is ensuring effective feature interaction between multiple modalities and bitemporality. Traditional approaches employ various techniques, such as addition \cite{2-RS3Mamba, 4-FreMamba, 15-CMS2I-Mamba, 16-DC-Mamba, 26-CVMH-UNet, 28-FusionMamba, 32-RemoteDet-Mamba, 41-MF-VMamba, 78-CSMN, 97-MGMF}, subtraction \cite{6-CDMamba, 8-IMDCD, 65-ConMamba, 91-SDMSPan}, direct concatenation along the $C$ dimension \cite{1-RS-Mamba, 22-Mamba-Diffusion, 35-AFA-Mamba, 41-MF-VMamba, 65-ConMamba, 84-COMO, 92-MambaDS}, CNN-based attention mechanisms \cite{23-MFMamba, 27-HLMamba, 35-AFA-Mamba, 36-MiM, 41-MF-VMamba, 52-SSRFN, 68-TTMGNet, 84-COMO, 92-MambaDS}, and cross-attention mechanism \cite{6-CDMamba, 34-MambaFormer, 91-SDMSPan} to facilitate multimodal and bi-temporal feature interactions.

Some studies have explored the potential of Mamba for multimodal and bi-temporal feature interaction. In this paper, we provide a comprehensive summary of all methods employed for multimodal and bi-temporal feature interaction in the remote sensing domain. We categorize these methods into four main groups:
(1) SSM formula-based methods \cite{84-COMO, 28-FusionMamba, 29-MSFMamba, 64-S2Mamba, 47-SegMamba-OS},
(2) Scan strategy-based feature preprocessing methods \cite{15-CMS2I-Mamba, 57-Pan-Mamba},
(3) Scan strategy-based feature post-processing methods \cite{14-ChangeMamba, 22-Mamba-Diffusion, 110-CD-Lamba, 113-TrackingMamba, 12-DMM}, and
(4) Mamba's gated mechanism-based methods \cite{48-RSCaMa, 57-Pan-Mamba}.

\begin{figure*}[h!]
\centering
\includegraphics[width=1\textwidth]{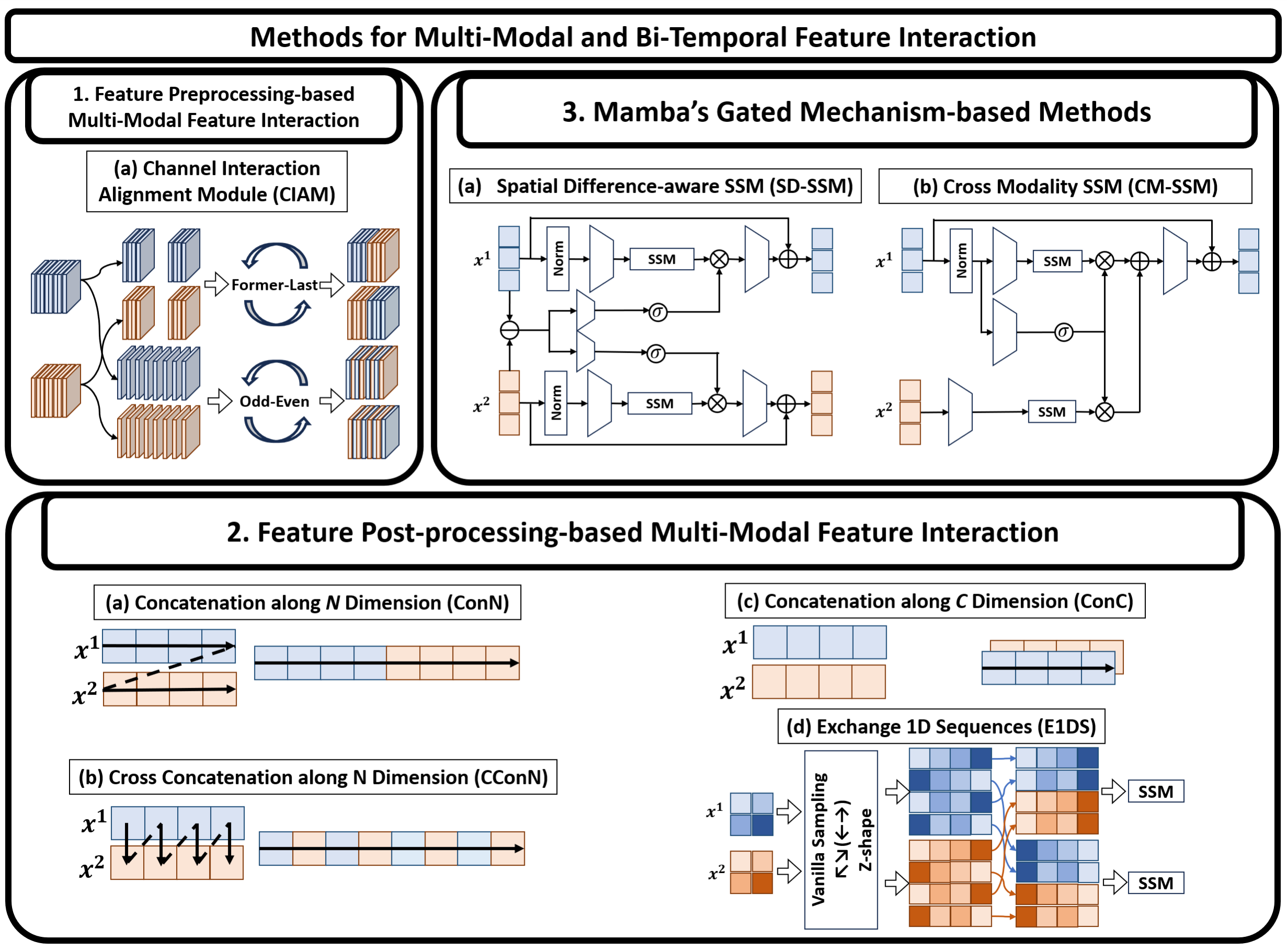}
\caption{Overview of multimodal and bi-temporal feature interaction methods. This figure illustrates the primary categories of multimodal and bi-temporal feature interaction approaches via Mamba, including (1) scan strategy's preprocessing-based methods, (2) scan strategy's post-processing-based methods, and (3) Mamba's gated mechanism-based methods.}
\label{fig:multimodality}
\end{figure*}

\subsubsection{\textbf{SSM Formula-based Methods}}
Standard SSMs are originally designed for single-modal and single-temporal features. Several studies have attempted to integrate multimodal and bi-temporal feature interactions within SSM formula, as discussed in Section \ref{sec:crossSSM} and illustrated in Fig. \ref{fig:SSM}.b.

\subsubsection{\textbf{Scan Strategy-based Feature Preprocessing Methods}} \label{sec:mulplePre}
Following the pre-processing methods of scan strategy framework introduced in Section \ref{sec:preprocess}, the Channel Interaction Alignment Module (CIAM) proposed in CMS2I-Mamba \cite{15-CMS2I-Mamba} can be categorized as a scan strategy's feature preprocessing method, as depicted in Fig. \ref{fig:multimodality}.1. CIAM splits each modality features into two separate pathways, employing Former-Last and Odd-Even concatenation strategies. This enables Mamba blocks to model features encompassing both modalities. Similarly, Pan-Mamba \cite{57-Pan-Mamba} only adopts the Former-Last strategy of CIAM to preprocess multimodal and bi-temporal features.

\subsubsection{\textbf{Scan Strategy-based Feature Post-Processing Methods}} \label{sec:mulplePost}
Similarly, following post-processing methods of the scan strategy framework in Section \ref{sec:post}, a total of four methods \cite{14-ChangeMamba, 22-Mamba-Diffusion, 110-CD-Lamba, 113-TrackingMamba, 12-DMM} fall under the category of the scan strategy's feature post-processing-based methods. Unlike preprocessing-based methods, these approaches are applied after obtaining 1D sequences.

The first method, Concatenation along the $N$ dimension (\textbf{ConN}) \cite{12-DMM, 14-ChangeMamba, 113-TrackingMamba}, directly concatenates multimodal and bi-temporal features along the $N$ dimension, as illustrated in Fig. \ref{fig:multimodality}.2a. ConN allows hidden states from the first modality to propagate into the computation of the second modality, thereby enabling SSMs to model cross-modal dependencies.
The second method, Cross Concatenation along the $N$ dimension (\textbf{CConN}) \cite{110-CD-Lamba, 14-ChangeMamba}, employs an interleaved concatenation strategy (Fig. \ref{fig:multimodality}.2b). Unlike ConN, which facilitates unidirectional interaction (from the first modality to the second), CConN enables bidirectional interaction, allowing both modalities to influence each other simultaneously.
The third method, Concatenation along the $C$ dimension (\textbf{ConC}) \cite{14-ChangeMamba, 48-RSCaMa}, merges multimodal and bi-temporal features along the $C$ dimension (Fig. \ref{fig:multimodality}.2c). This approach allows SSMs to capture long-range dependencies across modalities in parallel rather than sequentially.
The fourth method, Exchange 1D Sequences (\textbf{E1DS}), proposed in Mamba-Diffusion \cite{22-Mamba-Diffusion}, enables an SSM to process identical orientations from both modalities simultaneously (Fig. \ref{fig:multimodality}.2d). E1DS ensures that the projection layer's weights are optimized for both modalities concurrently during training.

\subsubsection{\textbf{Mamba's Gated Mechanism-based Methods}}
Beyond the methods discussed above, the fourth category explores Mamba’s gating mechanism to facilitate multimodal and bi-temporal feature interaction. RSCaMa \cite{48-RSCaMa} presents the Spatial Difference-aware SSM (SD-SSM), which replaces the traditional single-modality input for computing the activation value $Z$ with the difference between features from two modalities. This design enables the gating mechanism to explicitly model cross-modal interactions. Pan-Mamba \cite{57-Pan-Mamba} introduces Cross Modality SSM (CM-SSM), where multimodal and bi-temporal features are first projected into a unified representation space. CM-SSM then applies gated operations to enhance the learning of complementary signals and mitigate redundancy, leading to more effective multimodal and bi-temporal interaction.

%% file: Sec/5_macro.tex
\section{Macro-Architecture Advancement} \label{sec:macro}
Building upon the micro-architecture advancements discussed previously, this section examines higher-level architectural designs. We systematically analyze 4 key architectural advancement: (1) Hybrid Architectures with CNNs/Transformers (Section \ref{sec:Hybrid}), exploring combinations of Mamba with CNNs and Transformers; (2) Substitution in Existing Frameworks (Section \ref{sec:subtition}), where Mamba blocks are placed in existing frameworks; (3) Learning Paradigms (Section \ref{sec:supervisedlearning}), investigating unsupervised, self-supervised and prompt learning paradigms; and (4) Frequency Domain Operations (Section \ref{sec:fourier}), examining Fast Fourier Transformer, 2D Discrete Cosine Transform and Wavelet Transform. Tab.~\ref{tab:macro} presents the outline and the representative work.

\begin{table}[]
\renewcommand{\arraystretch}{1.38} 
\caption{Summary of macro-architecture advancements in remote sensing leveraging Mamba-based models, categorized into hybrid architectures with CNNs/Transformers, substitutions of established frameworks (U-Net \cite{UNet}, YOLO \cite{YOLO} and Diffusion Models \cite{diffusion}), learning paradigms, and frequency domain operations, along with corresponding representative studies.} 
\resizebox{\linewidth}{!}{ 
\begin{tabular}{cccl}
\hline
\multicolumn{4}{l}{\cellcolor[HTML]{C0C0C0}\textbf{Section \ref{sec:Hybrid}: Hybrid Architectures with CNNs/Transformer}}   \\ \hline

\multicolumn{1}{c|}{} & \multicolumn{1}{c|}{}  & \multicolumn{1}{c|}{Seq}  
        & \begin{tabular}[c]{@{}l@{}@{}@{}} 
            \cite{16-DC-Mamba},   \cite{19-Mamba-CR},  \cite{30-Res-Mamba}, \cite{53-ConvMambaSR},      \cite{92-MambaDS},  \cite{44-UVMSR}   \\ 
            \cite{27-HLMamba},      \cite{34-MambaFormer},\cite{40-PPMamba},   \cite{45-DualMamba}, \cite{71-LS2SM-MHMambaOut}, \cite{46-LE-Mamba}, \cite{78-CSMN}    \\ 
            \cite{38-MambaLG},      \cite{56-PPMamba},    \cite{102-LCCDMamba},\cite{83-MTIE-Net},  \cite{69-EGCM-UNet},        \cite{46-LE-Mamba}, \cite{38-MambaLG} \\
            \cite{56-PPMamba},      \cite{78-CSMN},        \cite{92-MambaDS},  \cite{93-HSIMamba},  \cite{111-HSRMamba},        \cite{63-GraphMamba},\\
            \cite{21-UV-Mamba},     \cite{52-SSRFN}
        \end{tabular}                                    \\ \cline{3-4} 

\multicolumn{1}{c|}{}  & \multicolumn{1}{c|}{\multirow{-2}{*}{Unit}} & \multicolumn{1}{c|}{Para} 
        & \cite{21-UV-Mamba}, \cite{8-IMDCD}, \cite{23-MFMamba}, \cite{40-PPMamba}, \cite{41-MF-VMamba}, \cite{99-MiM-ISTD}, \cite{102-LCCDMamba} \\ \cline{2-4} 

\multicolumn{1}{c|}{} & \multicolumn{1}{c|}{}  & \multicolumn{1}{c|}{Seq}  
        & \begin{tabular}[c]{@{}l@{}@{}} 
            \cite{2-RS3Mamba},        \cite{8-IMDCD},         \cite{23-MFMamba},  \cite{40-PPMamba},  \cite{41-MF-VMamba},    \cite{99-MiM-ISTD},     \cite{102-LCCDMamba}\\
            \cite{25-CM-UNet},        \cite{42-PyramidMamba}, \cite{72-RFCC},     \cite{96-MaDiNet},  \cite{3-RTMamba},       \cite{5-S2CrossMamba} \\ 
             \cite{33-SpectralMamba}, \cite{54-DBMamba},       \cite{59-STMamba}, \cite{25-CM-UNet},  \cite{42-PyramidMamba}, \cite{72-RFCC},         \cite{96-MaDiNet}
        \end{tabular}     \\ \cline{3-4} 
\multicolumn{1}{c|}{\multirow{-4}{*}{CNN}}     & \multicolumn{1}{c|}{\multirow{-2}{*}{Arch}} & \multicolumn{1}{c|}{Para} 
        & \begin{tabular}[c]{@{}l@{}} 
            \cite{53-ConvMambaSR}, \cite{65-ConMamba}, \cite{86-Prompt-Mamba}, \cite{89-HSIRMamba}, \cite{103-SCMamba}\\
             \cite{104-HMCNet}, \cite{106-Weaba}, \cite{2-RS3Mamba}, \cite{81-ColorMamba}
        \end{tabular} \\ \hline
        
\multicolumn{1}{c|}{}  & \multicolumn{1}{c|}{} & \multicolumn{1}{c|}{Seq}  
        &  \cite{17-RSMamba}, \cite{19-Mamba-CR},\cite{43-MorpMamba}, \cite{77-MHSSMamba}, \cite{91-SDMSPan}  \\ \cline{3-4} 
\multicolumn{1}{c|}{}  & \multicolumn{1}{c|}{\multirow{-2}{*}{Unit}} & \multicolumn{1}{c|}{Para} 
        & \cite{23-MFMamba}, \cite{34-MambaFormer}, \cite{70-TransMamba} \\ \cline{2-4} 
\multicolumn{1}{c|}{}  & \multicolumn{1}{c|}{}  & \multicolumn{1}{c|}{Seq}  
        & \begin{tabular}[c]{@{}l@{}} 
            \cite{39-MSTFNet}, \cite{42-PyramidMamba}, \cite{18-UNetMamba}, \cite{96-MaDiNet}   \\  
            \cite{48-RSCaMa}, \cite{105-Mamba-MDRNet} 
        \end{tabular}  \\ \cline{3-4} 
        
\multicolumn{1}{c|}{\multirow{-4}{*}{\begin{tabular}[c]{@{}c@{}}Trans-\\ former\end{tabular}}} & \multicolumn{1}{c|}{\multirow{-2}{*}{Arch}} 
    & \multicolumn{1}{c|}{Para} & \cite{76-MamTrans}    \\ \hline
\multicolumn{3}{c|}{Mamba as Module}  
    & \begin{tabular}[c]{@{}l@{}} 
         \cite{6-CDMamba}, \cite{12-DMM}, \cite{32-RemoteDet-Mamba}, \cite{47-SegMamba-OS}, \cite{84-COMO}, \cite{110-CD-Lamba}  \\
         \cite{37-HSDet-Mamba}, \cite{87-Mamba-UAV-SegNet}, \cite{75-ES-HS-FPN}  \\ \hline
    \end{tabular}   \\ \hline

\multicolumn{4}{l}{\cellcolor[HTML]{C0C0C0}\textbf{Section \ref{sec:subtition}: Substitution in Existing Framework}}  \\ \hline
\multicolumn{3}{c|}{U-Net}      
    & \begin{tabular}[c]{@{}l@{}@{}@{}} 
        \cite{2-RS3Mamba}, \cite{7-RSDehamba}, \cite{14-ChangeMamba}, \cite{16-DC-Mamba}, \cite{17-RSMamba}, \cite{18-UNetMamba}  \\ 
        \cite{19-Mamba-CR}, \cite{21-UV-Mamba}, \cite{23-MFMamba}, \cite{24-LightMamba}, \cite{25-CM-UNet}, \cite{26-CVMH-UNet}\\
        \cite{40-PPMamba}, \cite{41-MF-VMamba}, \cite{44-UVMSR}, \cite{56-PPMamba}, \cite{66-DHM}, \cite{69-EGCM-UNet}, \cite{70-TransMamba}  \\ 
        \cite{72-RFCC}, \cite{79-SSUMamba}, \cite{80-FMambaIR}, \cite{81-ColorMamba}, \cite{83-MTIE-Net}, \cite{98-LGMamba}, \cite{99-MiM-ISTD} \\
        \cite{102-LCCDMamba}, \cite{104-HMCNet}, \cite{106-Weaba}, \cite{109-UrbanSSF} \\    
    \end{tabular}   \\ \hline
\multicolumn{3}{c|}{YOLO}       & \cite{37-HSDet-Mamba}, \cite{75-ES-HS-FPN}, \cite{85-YOLO-Mamba}, \cite{94-HRMamba-YOLO} \\ \hline
\multicolumn{3}{c|}{Diffusion}  & \cite{8-IMDCD}, \cite{96-MaDiNet}                                          \\ \hline

\multicolumn{4}{l}{\cellcolor[HTML]{C0C0C0}\textbf{Section \ref{sec:supervisedlearning}: Learning Paradigm}}   \\ \hline
\multicolumn{3}{c|}{Unsupervised Learning}      &  \cite{72-RFCC}                       \\ \hline
\multicolumn{3}{c|}{Self-Learning}      &  \cite{73-HTD-Mamba}, \cite{108-SatMamba}    \\ \hline
\multicolumn{3}{c|}{Prompt Learning}   &  \cite{86-Prompt-Mamba}               \\ \hline

\multicolumn{4}{l}{\cellcolor[HTML]{C0C0C0}\textbf{Section \ref{sec:fourier}: Frequency Domain Operation}}  \\ \hline
\multicolumn{3}{c|}{Fourier Transform}            &  \cite{4-FreMamba}, \cite{34-MambaFormer}, \cite{70-TransMamba}, \cite{78-CSMN}, \cite{80-FMambaIR}, \cite{88-VMambaSCI} \\ \hline
\multicolumn{3}{c|}{Wavelet Transform}            &  \cite{61-WaveMamba} \\ \hline
\multicolumn{3}{c|}{ \begin{tabular}[c]{@{}l@{}}
    2D Discrete \\ Cosine Transform \end{tabular}} & \cite{26-CVMH-UNet} \\ \hline

\end{tabular}
}
\label{tab:macro}
\end{table}

\subsection{Hybrid Architectures with CNNs/Transformer} \label{sec:Hybrid}

\subsubsection{\textbf{Hybrid Architecture with CNNs}}
The integration of CNNs with Mamba aims to complement the global modeling capability of Mamba with the local feature extraction capability of CNNs. Various approaches \cite{2-RS3Mamba, 3-RTMamba,5-S2CrossMamba, 6-CDMamba, 8-IMDCD, 12-DMM, 16-DC-Mamba, 18-UNetMamba, 19-Mamba-CR, 21-UV-Mamba, 23-MFMamba, 25-CM-UNet, 26-CVMH-UNet, 27-HLMamba, 30-Res-Mamba, 32-RemoteDet-Mamba, 33-SpectralMamba, 34-MambaFormer,35-AFA-Mamba, 36-MiM, 37-HSDet-Mamba, 38-MambaLG, 40-PPMamba, 41-MF-VMamba, 42-PyramidMamba, 44-UVMSR, 45-DualMamba, 46-LE-Mamba, 47-SegMamba-OS, 52-SSRFN, 53-ConvMambaSR, 54-DBMamba, 56-PPMamba, 59-STMamba, 63-GraphMamba, 65-ConMamba, 68-TTMGNet, 69-EGCM-UNet, 71-LS2SM-MHMambaOut, 72-RFCC, 74-SITSMamba, 75-ES-HS-FPN, 78-CSMN, 79-SSUMamba, 81-ColorMamba, 82-HDMba, 84-COMO, 86-Prompt-Mamba, 87-Mamba-UAV-SegNet, 88-VMambaSCI, 89-HSIRMamba, 92-MambaDS, 93-HSIMamba, 94-HRMamba-YOLO, 95-LDMNet, 96-MaDiNet, 97-MGMF, 99-MiM-ISTD, 102-LCCDMamba, 103-SCMamba, 104-HMCNet, 106-Weaba, 109-UrbanSSF, 111-HSRMamba} have been proposed to achieve this hybridization, typically falling into sequential or parallel configurations—both within each \textbf{stackable} building block and across the overall \textbf{architectural} framework. 

\paragraph{Sequential or Parallel Integration within Basic Units}
One common approach to integrate CNNs with Mamba is in a sequential manner, wherein CNN layers may be potentially combined with normalization layers and activation functions \cite{16-DC-Mamba, 19-Mamba-CR, 30-Res-Mamba, 53-ConvMambaSR, 92-MambaDS, 44-UVMSR, 27-HLMamba, 34-MambaFormer, 40-PPMamba, 45-DualMamba, 71-LS2SM-MHMambaOut}. Alternatively, CNNs can be arranged in parallel with Mamba within a basic unit, facilitating complementary feature extraction \cite{21-UV-Mamba, 8-IMDCD, 23-MFMamba, 40-PPMamba, 41-MF-VMamba, 99-MiM-ISTD, 102-LCCDMamba}.

Beyond simple CNN-Mamba combinations, several studies incorporate multi-scale CNNs to enhance local feature extraction in a sequential manner \cite{46-LE-Mamba, 78-CSMN, 38-MambaLG, 56-PPMamba, 102-LCCDMamba, 83-MTIE-Net}. Similarly, CNN-based attention mechanisms have been employed to refine feature representations when combined sequentially with Mamba \cite{46-LE-Mamba, 38-MambaLG, 56-PPMamba, 78-CSMN, 92-MambaDS, 93-HSIMamba, 111-HSRMamba}.

In addition to these designs, other approaches incorporate specialized CNN variations to enhance performance. GraphMamba \cite{63-GraphMamba} integrates Graph CNNs (GCN) with Mamba sequentially, while UV-Mamba \cite{21-UV-Mamba} utilizes Deformable CNNv4 (DCNv4) \cite{DCN, DCNv2} sequentially. SSRFN \cite{52-SSRFN} employs 3D CNNs in parallel with Mamba to preserve spatial information. EGCM-UNet \cite{69-EGCM-UNet} enhances edge feature extraction by incorporating CNNs with $3\times1$ and $1\times3$ kernels.

\paragraph{Sequential or Parallel Integration at the Architectural Level}
Beyond modifications at the unit level, hybrid architectures combining CNNs and Mamba have also been explored at the architecture level. For segmentation tasks, some architectures adopt a Mamba-based encoder paired with a CNN-based decoder \cite{2-RS3Mamba, 8-IMDCD, 23-MFMamba, 40-PPMamba, 41-MF-VMamba, 99-MiM-ISTD, 102-LCCDMamba}, while others employ a CNN-based encoder with a Mamba-based decoder \cite{25-CM-UNet, 42-PyramidMamba, 72-RFCC, 96-MaDiNet}.
For the non-encoder-decoder architectures, hybrid approaches often distribute CNNs and Mamba across different processing stages. Some models apply CNNs in the first several stages, followed by Mamba in later stages \cite{3-RTMamba, 5-S2CrossMamba, 33-SpectralMamba, 54-DBMamba, 59-STMamba}. Conversely, others adopt the opposite configuration, employing Mamba in the early stages and CNNs in the later stages \cite{25-CM-UNet, 42-PyramidMamba, 72-RFCC, 96-MaDiNet}. 

Apart from such sequential structures, hybrid architectures utilizing parallel structures have been proposed to simultaneously capture both local and global features. Some models establish two separate branches—one Mamba-based and the other CNN-based—before fusing the extracted global and local features \cite{53-ConvMambaSR, 65-ConMamba, 86-Prompt-Mamba, 89-HSIRMamba, 103-SCMamba, 104-HMCNet, 106-Weaba}. RS3Mamba \cite{2-RS3Mamba} and ColorMamba \cite{81-ColorMamba} introduce Mamba-based auxiliary encoders to provide global information to CNN-based main encoders, facilitating feature fusion at each processing stage.

\subsubsection{\textbf{Hybrid Architecture with Transformers}}

Efforts to enhance Mamba's sequential global modeling with Transformer's parallel global modeling have led to various Transformer-Mamba hybrid architectures. Similar to the CNN-Mamba hybrid approaches, these Transformer-Mamba architectures can be categorized into sequential or parallel structures within each \textbf{stackable} basic unit and at the \textbf{architectural} level.

\paragraph{Sequential or Parallel Integration within Basic Units}
Some studies simply integrate vanilla Transformer blocks with Mamba sequentially \cite{17-RSMamba, 19-Mamba-CR, 43-MorpMamba, 77-MHSSMamba, 91-SDMSPan}. In addition to these sequential integration methods, SDMSPan \cite{91-SDMSPan} employs cross-attention to facilitate multimodal and bi-temporal feature interaction at the beginning of each basic unit.

Furthermore, MFMamba \cite{23-MFMamba} and MambaFormerSR \cite{34-MambaFormer} integrate vanilla Transformer blocks alongside Mamba in a parallel configuration. TransMamba \cite{70-TransMamba} improves spectral-domain representation by incorporating a Transformer block augmented with Fourier transform operations.


\paragraph{Sequential or Parallel Integration at the Architectural Level}
At the architectural level, MamTrans \cite{76-MamTrans} is the only work that integrates Transformer and Mamba in parallel, whereas other studies adopt a sequential design.
MSTFNet \cite{39-MSTFNet} places Mamba blocks in the early stages and Swin-Transformer \cite{SwinTransformer} in the later stages. PyramidMamba \cite{42-PyramidMamba}, UNetMamba \cite{18-UNetMamba}, and MaDiNet \cite{96-MaDiNet} utilize vanilla Transformer, efficient Transformer \cite{EfficientTransformer}, and agent Transformer \cite{agenttransformer} encoders, respectively, paired with Mamba-based decoders. RSCaMa \cite{48-RSCaMa} and Mamba-MDRNet \cite{105-Mamba-MDRNet} initially extract features using a Mamba-based backbone, which are then fed into a Transformer-based language model to produce text captions and predict classification outputs.

\subsubsection{\textbf{Mamba as a Module}}
Beyond serving as a backbone, Mamba has also been employed as a module to achieve specific objectives. Many studies \cite{6-CDMamba, 12-DMM, 32-RemoteDet-Mamba, 47-SegMamba-OS, 84-COMO, 110-CD-Lamba} utilize Mamba-based modules to facilitate multi-modal, bi-temporal, multi-scale, and multi-stage feature interactions and fusion.
In particular, HSDet-Mamba \cite{37-HSDet-Mamba} and Mamba-UAV-SegNet \cite{87-Mamba-UAV-SegNet} incorporate Mamba-based modules to enhance global multi-scale feature representation. Additionally, ES-HS-FPN \cite{75-ES-HS-FPN} employs Mamba to aggregate global features with local object features, further improving feature integration.

\subsection{Substitution in Existing Frameworks} \label{sec:subtition}

\subsubsection{\textbf{Substitution in U-Net}}
U-Net \cite{UNet} is a symmetric encoder-decoder architecture extensively utilized in image segmentation. It is distinguished by its skip connections, which integrate shallow spatial details with deep semantic features across hierarchical scales, thereby facilitating precise localization and multi-scale contextual modeling. Due to the high efficacy of the U-Net framework, several studies \cite{7-RSDehamba, 17-RSMamba, 18-UNetMamba, 19-Mamba-CR, 21-UV-Mamba, 24-LightMamba, 40-PPMamba, 44-UVMSR, 56-PPMamba, 66-DHM, 72-RFCC, 79-SSUMamba, 80-FMambaIR, 83-MTIE-Net, 98-LGMamba, 99-MiM-ISTD, 104-HMCNet, 106-Weaba} have replaced the conventional CNN-based or Transformer-based blocks with Mamba-based blocks to enhance feature representation and computational efficiency. Furthermore, some approaches \cite{14-ChangeMamba, 16-DC-Mamba, 41-MF-VMamba, 70-TransMamba} have introduced a Siamese encoder structure to effectively process multi-modal and bi-temporal images.

Building upon the original U-Net framework, some studies \cite{2-RS3Mamba, 69-EGCM-UNet, 81-ColorMamba, 23-MFMamba} have designed auxiliary branches to augment feature extraction capabilities. For instance, RS3Mamba \cite{2-RS3Mamba} incorporates a Mamba-based auxiliary encoder to supplement global information extracted by a CNN-based encoder, thereby enhancing local feature representation. EGCM-UNet \cite{69-EGCM-UNet} introduces an auxiliary branch that extracts edge features, which are subsequently fused at each stage of the encoder-decoder pipeline. ColorMamba \cite{81-ColorMamba} integrates a Mamba-based HSV color prediction sub-network, leveraging HSV color space priors to provide multi-scale guidance for RGB reconstruction. Similarly, MFMamba~\cite{23-MFMamba} utilizes a Mamba-based auxiliary encoder to extract global features from digital surface model (DSM) images, which are then fused with local features from RGB images, leading to improved overall performance.

Additionally, several studies have sought to refine feature representation within skip connections. Traditional approaches include leveraging multi-scale CNNs and attention mechanisms \cite{25-CM-UNet}, introducing frequency domain terms \cite{26-CVMH-UNet}, and incorporating self-attention mechanisms \cite{109-UrbanSSF}. Beyond these methods, Mamba-based blocks have been employed to enhance skip connections. 
LCCDMamba \cite{102-LCCDMamba} first extracts local features using multi-scale CNNs and subsequently applies a Mamba-based block for long-range feature modeling.

\subsubsection{\textbf{Substitution in YOLO}}
To enhance object detection performance, several studies have explored the integration of Mamba-based modules into the YOLO \cite{YOLO} framework. These modifications primarily target the backbone and feature fusion components of YOLOv8 \cite{yolov8} to improve global feature extraction while maintaining computational efficiency. HSDet-Mamba \cite{37-HSDet-Mamba} integrates Mamba into the YOLO architecture by replacing conventional feature extraction modules with a spatial feature enhancement module (SFEM). This module fuses CNN-based feature extraction with Mamba to better capture both spatial and spectral dependencies, leading to improved detection accuracy in hyperspectral imagery.
ES-HS-FPN \cite{75-ES-HS-FPN} replaces the Spatial Pyramid Pooling-Fast (SPPF) module of YOLOv8 with an SPPF-Mamba module, which enhances global and local feature fusion, thereby improving object-context representation. 
In YOLO-Mamba \cite{85-YOLO-Mamba}, the C2f module in YOLOv8 is replaced with a C2f-Mamba module, aiming at leveraging a Mamba-based attention mechanism to capture long-range dependencies across feature and spatial dimensions, reducing redundant information and improving the detection of small or occluded objects in infrared aerial imagery. The resulting YOLO-Mamba model achieves higher detection accuracy while maintaining minimal computational overhead.
HRMamba-YOLO \cite{94-HRMamba-YOLO} integrates Mamba into the YOLO architecture,
which enhances feature extraction and multi-scale feature fusion by capturing long-range dependencies and improving contextual information representation. The incorporation of Mamba-based modules within the high-resolution feature pyramid network further strengthens cross-scale feature interactions, leading to improved small object detection performance in UAV imagery.

\subsubsection{\textbf{Substitution in Diffusion Model}}
Several studies have explored the integration of Mamba-based modules into diffusion models \cite{diffusion} to enhance feature extraction and representation learning in remote sensing tasks. In particular, MaDiNet \cite{96-MaDiNet} replaces conventional CNN-based feature extraction in diffusion-based SAR target detection with the MambaSAR module, which captures rich spatial structural information and improves target differentiation from complex backgrounds. This integration allows MaDiNet to enhance global contextual understanding while leveraging the denoising process of the diffusion model to refine target localization.
IMDCD \cite{8-IMDCD} incorporates Mamba into the diffusion model through the Swin-Mamba-Encoder (SME) and the Variable State Space Change Detection (VSS-CD) module. SME enhances long-range dependency modeling, while VSS-CD extracts transformation-aware features, which are iteratively refined within the diffusion process to generate high-precision change detection maps. The diffusion model provides a generative framework for iterative refinement, ensuring robustness against noise and improving detection accuracy.

\subsection{Learning Paradigm } \label{sec:supervisedlearning}
This subsection provides an overview of various learning paradigms applied in remote sensing, with emphasis on unsupervised, self-supervised and prompt learning paradigms. Except for these paradigms, self-supervised learning is used for the rest of the work.

\subsubsection{\textbf{Unsupervised Learning}}
In the reviewed study, only RFCC \cite{72-RFCC} employs an unsupervised learning approach for remote sensing image change detection. The method integrates Mamba-based differentiable feature clustering to perform automatic segmentation, ensuring that spatially contiguous pixels with similar spectral and spatial features are grouped together. To further enhance classification accuracy, the framework incorporates fuzzy C-means clustering, which decomposes mixed pixels into multiple signal classes, and a context-sensitive Bayesian network (CSBN) \cite{bayesian}, which refines posterior probability estimations by incorporating spatial information. This combination reduces the cumulative clustering error and eliminates the reliance on manual annotations.

\subsubsection{\textbf{Self-Supervised Learning}}
In the reviewed studies, two innovative frameworks utilizing Mamba exhibit distinct self-supervised strategies. In particular, SatMamba \cite{108-SatMamba} integrates masked autoencoders (MAEs) \cite{MAE} with Mamba for foundation model pretraining. Inspired by MAE \cite{MAE} and SatMAE \cite{SatMAE}, the framework randomly masks 75\% of image patches during pretraining on the fMoW dataset \cite{fMoW} and reconstructs normalized pixel values of masked regions through a Mamba-based encoder-decoder structure. This approach eliminates dependency on labeled data while capturing spatial-spectral dependencies through linear-complexity SSM blocks. Notably, positional encodings are experimentally ablated, revealing that Mamba's inherent sequential processing effectively preserves spatial order without explicit positional guidance. HTD-Mamba \cite{73-HTD-Mamba} introduces spectrally contrastive learning for hyperspectral target detection. It employs a spatial-encoded spectral augmentation technique to generate augmented views by aggregating contextual pixels within patches, weighted by spectral similarity. These views form positive/negative pairs for contrastive loss optimization, enabling discrimination between target and background spectra without manual annotations. 

\subsubsection{\textbf{Prompt Learning}}
Prompt-Mamba \cite{86-Prompt-Mamba} introduces an interactive prompt-based segmentation framework for urban flood detection, leveraging four distinct types of prompts (i.e., points, boxes, curves, and masks), to guide the model in refining segmentation results. The method employs a convolutional prompt encoder that transforms prompt inputs into structured feature representations, enabling efficient integration with image embeddings. By incorporating expert knowledge through interactive prompts, the approach effectively reduces annotation costs while maintaining segmentation accuracy.

\subsection{Frequency Domain Operation}  \label{sec:fourier}
Several studies incorporate frequency-domain operations, namely Fast Fourier Transform (FFT), wavelet transform and 2D discrete cosine transform, to enrich spatial-spectral features with complementary frequency information. In total, seven papers \cite{4-FreMamba, 34-MambaFormer, 70-TransMamba, 78-CSMN, 80-FMambaIR, 88-VMambaSCI} integrate FFT, while one paper \cite{61-WaveMamba} employs wavelet transformation and one paper \cite{26-CVMH-UNet} uses 2D discrete cosine transform.

\subsubsection{\textbf{Fast Fourier Transform}}
FreMamba \cite{4-FreMamba} is the first to merge Mamba with the Fourier transformation within its basic unit, adding an FFT-based branch sequentially to both the Mamba-based and CNN-based blocks. This design facilitates the exploration of spatial and frequency correlations.  Similarly, FMambaIR \cite{80-FMambaIR} applies an attention mechanism separately to the amplitude and phase obtained via FFT, alongside an additional Mamba branch, thereby enhancing the extraction of global degradation features and overall global information perception. MambaFormerSR \cite{34-MambaFormer} introduces an FFT-based attention mechanism into the conventional Feed-Forward Network (FFN). By multiplying the FFT-enhanced branch (processed through CNNs and FFT) with a branch solely processed by CNNs, the network is guided to emphasize degradation-sensitive components during image restoration. In TransMamba \cite{70-TransMamba}, a dual-branch architecture comprised Transformer and Mamba modules is proposed. Within the Transformer branch, a self-attention module applies FFT to reallocate features into distinct frequency bands. Learnable attention weights then adaptively suppress low-frequency rain streaks while amplifying high-frequency textures, and a Spectral Enhanced Feed-Forward (SEFF) module further refines features with frequency-specific filters and dilated convolutions.

In addition to integration at the basic unit level, two other studies incorporate FFT in different network components. CSMN \cite{78-CSMN} proposes a Cross-Domain Mamba Module (CDMM) that fuses spatial, spectral, and frequency data by applying a Fourier transform after convolutional feature fusion. This branch captures global frequency patterns, contributing to improved detail preservation and robust spatial-spectral fusion in pan-sharpening tasks. Furthermore, VmambaSCI \cite{88-VMambaSCI} integrates FFT within multi-stage interactions to reinforce feature fusion across different stages.

\subsubsection{\textbf{2D Discrete Cosine Transform}}
CVMH-UNet \cite{26-CVMH-UNet} employs a Multi-Frequency Multi-Scale Feature Fusion Block (MFMSBlock) that uses a 2D Discrete Cosine Transform (DCT) to compute channel attention. This allows the model to better capture both low-frequency structural information and high-frequency edge details, mitigating information loss in skip connections and thereby enhancing segmentation accuracy. This approach not only enhances feature fusion across multiple scales but also addresses the issue of information inconsistency in conventional U-Net skip connections.

\subsubsection{\textbf{Wavelet Transform}}
WaveMamba \cite{61-WaveMamba} integrates wavelet transform with a Mamba-based spatial–spectral network by employing the classical Haar wavelet to decompose hyperspectral data into multiple subbands. This approach separates input spatial and spectral features into low- and high-frequency components, effectively capturing both fine-grained details and global structures, and subsequently feeds the multi-resolution subband features into a SSM to boost classification performance.

%% file: Sec/6_downstream.tex
\section{Downstream Applications} \label{sec:downstramtask}

\begin{table}[t!]
\caption{The overview of five tasks in remote sensing and their representative work, including classification, semantic segmentation, object detection, change detection, super-resolution and image restoration.}
\resizebox{\linewidth}{!}{ 
\begin{tabular}{llll} \hline
Task & \multicolumn{3}{l}{Representative Works} \\ \hline
Classification & 
    \multicolumn{3}{l}{
    \begin{tabular}[l]{@{}l@{}@{}@{}@{}} 
        \cite{29-MSFMamba}, \cite{35-AFA-Mamba}, \cite{5-S2CrossMamba}, \cite{112-RSVMamba}, \cite{33-SpectralMamba}, \cite{39-MSTFNet}, \cite{43-MorpMamba}, \cite{45-DualMamba}, \cite{49-SS-Mamba}  \\
        \cite{61-WaveMamba}, \cite{63-GraphMamba}, \cite{74-SITSMamba}, \cite{89-HSIRMamba}, \cite{101-MambaHSI}, \cite{103-SCMamba}, \cite{15-CMS2I-Mamba}\\
        \cite{31-HyperMamba}, \cite{64-S2Mamba}, \cite{67-DTAM}, \cite{76-MamTrans},  \cite{93-HSIMamba}, \cite{107-MLMamba}, \cite{46-LE-Mamba}\\
        \cite{38-MambaLG}, \cite{60-SSUM}, \cite{59-STMamba}, \cite{36-MiM}, \cite{9-RSMamba}, \cite{55-3DSS-Mamba}, \cite{51-IGroupSS-Mamba}\\
         \cite{71-LS2SM-MHMambaOut}, \cite{54-DBMamba}, \cite{77-MHSSMamba}, \cite{10-rethinking},  \cite{50-G-VMamba},  
    \end{tabular}}   \\ \hline 
    
\begin{tabular}[l]{@{}l@{}} Semantic \\ Segmentation \end{tabular}
    & \multicolumn{3}{l}{ 
    \begin{tabular}[l]{@{}l@{}@{}@{}@{}} 
         \cite{1-RS-Mamba}, \cite{2-RS3Mamba}, \cite{3-RTMamba}, \cite{47-SegMamba-OS}, \cite{27-HLMamba}, \cite{13-Samba}, \cite{42-PyramidMamba}, \cite{95-LDMNet}\\
         \cite{26-CVMH-UNet}, \cite{18-UNetMamba}, \cite{22-Mamba-Diffusion}, \cite{23-MFMamba}, \cite{25-CM-UNet}, \cite{87-Mamba-UAV-SegNet}, \cite{105-Mamba-MDRNet} \\
         \cite{108-SatMamba}, \cite{40-PPMamba},  \cite{69-EGCM-UNet}, \cite{21-UV-Mamba}, \cite{56-PPMamba}, \cite{98-LGMamba}, \ \cite{109-UrbanSSF}\\
    \end{tabular}}   \\ \hline

\begin{tabular}[l]{@{}l@{}} Object \\ Detection \end{tabular}
    & \multicolumn{3}{l}{ 
    \begin{tabular}[c]{@{}l@{}@{}@{}@{}} 
         \cite{12-DMM}, \cite{32-RemoteDet-Mamba}, \cite{37-HSDet-Mamba}, \cite{84-COMO}, \cite{75-ES-HS-FPN}, \cite{85-YOLO-Mamba}, \cite{97-MGMF}\\
         \cite{96-MaDiNet}, \cite{94-HRMamba-YOLO}, \cite{100-YOLOv10SR}   \\
    \end{tabular}}   \\ \hline


\begin{tabular}[l]{@{}l@{}} Change \\ Detection \end{tabular}
    & \multicolumn{3}{l}{ 
    \begin{tabular}[c]{@{}l@{}@{}@{}@{}} 
        \cite{1-RS-Mamba}, \cite{8-IMDCD}, \cite{6-CDMamba}, \cite{68-TTMGNet}, \cite{14-ChangeMamba}, \cite{16-DC-Mamba}, \cite{102-LCCDMamba}, \cite{72-RFCC}      \\
        \cite{41-MF-VMamba}, \cite{110-CD-Lamba}, \cite{65-ConMamba}  \\
    \end{tabular}}   \\ \hline

\begin{tabular}[l]{@{}l@{}} Super- \\ Resolution \end{tabular}
    & \multicolumn{3}{l}{ 
    \begin{tabular}[c]{@{}l@{}@{}@{}@{}} 
         \cite{4-FreMamba}, \cite{52-SSRFN}, \cite{53-ConvMambaSR}, \cite{111-HSRMamba}, \cite{30-Res-Mamba}, \cite{34-MambaFormer}, \cite{44-UVMSR},  \cite{90-MaIR} \\
    \end{tabular}}   \\ \hline

\begin{tabular}[l]{@{}l@{}} Image \\ Restoration \end{tabular}
    & \multicolumn{3}{l}{ 
    \begin{tabular}[c]{@{}l@{}@{}@{}@{}} 
         \cite{7-RSDehamba},  \cite{17-RSMamba}, \cite{19-Mamba-CR}, \cite{82-HDMba}, \cite{80-FMambaIR}, \cite{24-LightMamba}, \cite{83-MTIE-Net} \\
         \cite{88-VMambaSCI}, \cite{79-SSUMamba}, \cite{106-Weaba}, \cite{70-TransMamba}, \cite{58-HSIDMamba}, \cite{66-DHM} \\         
    \end{tabular}}   \\ \hline





\hline
\end{tabular}
}
\label{tab:summarypaper}
\end{table}

\begin{table}[h]
\caption{The \textbf{classification} benchmark for the \textbf{Pavia University} Dataset \cite{pu}. \textbf{Rate} represents the percentage of the training data to the total data; \textbf{OA} represents overall classification accuracy; \textbf{AA} represents average classification accuracy; \textbf{Kappa} represents the kappa coefficient.}
\resizebox{\linewidth}{!}{%
\begin{tabular}{l|cccc}
\hline
\textbf{Methods} & \textbf{Rate} & \textbf{OA} & \textbf{AA} & \textbf{Kappa} \\ \hline
\multicolumn{5}{c}{\textbf{CNN-based Methods}} \\ \hline
1D-CNN \cite{cla_1DConv}  & 5.00\%  & 85.82 & 83.25 & 81.17 \\
2D-CNN \cite{cla_2DConv}  & 5.00\%  & 93.30 & 89.49 & 91.07 \\
3D-CNN \cite{cla_3DConv}  & 5.00\%  & 93.52 & 91.22 & 91.37 \\
1D-CNN \cite{cla_1D2}     & 9.90\%  & 75.50 & 86.26 & 69.48 \\
2D-CNN \cite{cla_2D2}     & 9.90\%  & 86.05 & 88.99 & 81.87 \\ \hline
\multicolumn{5}{c}{\textbf{Transformer-based Methods}} \\ \hline
HIS-BERT \cite{cla_BERT}           & 5.00\%  & 85.45 & 71.41 & 83.36 \\
GSC-ViT \cite{cla_GSC}             & 5.00\%  & 98.28 & 94.42 & 98.04 \\
CASST \cite{cla_CASST}             & 5.00\%  & 96.65 & 92.25 & 96.18 \\
ViT \cite{cla_vit}               & 9.90\%  & 76.99 & 80.22 & 70.10 \\
SpectralFormer \cite{cla_spectralFormer} & 9.90\%  & 91.07 & 90.20 & 88.05 \\
SSFTT \cite{cla_SSFTT}            & 9.90\%  & 92.61 & 93.37 & 90.29 \\ \hline
\multicolumn{5}{c}{\textbf{Mamba-based Methods}} \\ \hline
LS2SM \cite{71-LS2SM-MHMambaOut}    & 0.10\%  & 98.83 & 98.18 & 98.45 \\
SDMamba \cite{103-SCMamba}           & 0.21\%  & 98.50 & 95.89 & 98.01 \\
STMamba \cite{59-STMamba}            & 0.50\%  & 97.03 & 94.27 & 96.05 \\
MambaHSI \cite{101-MambaHSI}          & 0.63\%  & 95.74 & 95.86 & 95.00 \\
SSUM \cite{60-SSUM}                  & 1.00\%  & 96.15 & 95.42 & 94.91 \\
LE-Mamba \cite{46-LE-Mamba}          & 2.00\%  & 99.63 & 99.43 & 99.51 \\
IGroupSS-Mamba \cite{51-IGroupSS-Mamba} & 5.00\%  & \textbf{99.75} & \textbf{99.46} & \textbf{99.66} \\
DBMamba \cite{54-DBMamba}           & 5.00\%  & 99.40 & 98.99 & 99.21 \\
3DSS-Mamb \cite{55-3DSS-Mamba}       & 5.00\%  & 98.48 & 97.56 & 97.98 \\
MiM \cite{36-MiM}                  & 8.90\%  & 91.58 & 92.76 & 89.83 \\
MamTrans \cite{76-MamTrans}         & 8.90\%  & 96.30 & 95.83 & 95.02 \\
MambaLG \cite{38-MambaLG}           & 8.92\%  & 95.66 & 95.91 & 94.19 \\
S2Mamba \cite{64-S2Mamba}           & 8.93\%  & 97.81 & 97.14 & 97.05 \\
HSIRMamba \cite{89-HSIRMamba}        & 9.90\%  & 99.20 & 99.21 & 98.95 \\
HSIMamba \cite{93-HSIMamba}         & 9.90\%  & 98.08 & 97.87 & 97.41 \\
MorpMamba \cite{43-MorpMamba}        & 20.00\% & 97.67 & 96.93 & 96.91 \\
WaveMamba \cite{61-WaveMamba}        & 25.00\% & 98.63 & 97.70 & 98.19 \\
MHSSMamba \cite{77-MHSSMamba}        & 25.00\% & 96.41 & 97.62 & 96.85 \\
SS-Mamba \cite{49-SS-Mamba}          & 49.89\% & 96.40 & 98.43 & 95.31 \\
DTAM \cite{67-DTAM}                  & 50.00\% & 84.03 & -     & 81.62 \\ \hline
\end{tabular}%
}
\label{tab:classification}
\end{table}

\begin{table}[h]
\centering
\caption{The \textbf{segmentation} benchmark for the \textbf{ISPRS Vaihingen} dataset \cite{vaihingen}. \textbf{mF1} represents mean F1 score; \textbf{mIOU} represents mean intersection over union; \textbf{OA} represents overall accuracy.}
\resizebox{0.85\linewidth}{!}{%
\begin{tabular}{l|ccc}
\hline
\textbf{Methods} & \textbf{mF1} & \textbf{mIOU} & \textbf{OA} \\ \hline
\multicolumn{4}{c}{\textbf{CNN-based Methods}} \\ \hline
DANet \cite{seg_DANet}        & 79.60 & 69.40 & 88.20 \\
ABCNet \cite{seg_ABCUNet}      & 89.50 & 81.30 & 90.70 \\
CMTFNet \cite{seg_CMTFNet}      & 87.37 & 78.06 & -     \\
BANet \cite{seg_BANet}         & 90.32 & 82.45 & 91.92 \\
MANet \cite{seg_MANet}         & 90.68 & 83.06 & 92.28 \\ \hline
\multicolumn{4}{c}{\textbf{Transformer-based Methods}} \\ \hline
HST-UNet \cite{seg_HSTUNet}         & 86.62 & 78.67 & -     \\
FTUNetformer \cite{seg_FTUNetformer} & 91.30 & 84.10 & 91.60 \\
DC-Swin \cite{seg_DCSwin}         & 90.71 & 83.08 & 92.30 \\
UNetFormer \cite{seg_unetformer}      & 90.59 & 82.93 & 92.21 \\
TransUNet \cite{seg_transunet}        & 92.86 & 87.15 & 91.56 \\ \hline
\multicolumn{4}{c}{\textbf{Mamba-based Methods}} \\ \hline
RS3Mamba \cite{2-RS3Mamba}     & 90.34 & 82.78 & -     \\
RTMamba \cite{3-RTMamba}       & 91.08 & 83.92 & 91.30 \\
Samba \cite{13-Samba}          & 84.23 & 73.56 & -     \\
UNetMamba \cite{18-UNetMamba}  & 90.95 & 83.47 & 92.51 \\
MFMamba \cite{23-MFMamba}      & 90.52 & 83.13 & 91.81 \\
CM-Unet \cite{25-CM-UNet}      & \textbf{92.01} & \textbf{85.48} & \textbf{93.81} \\
CVMH-UNet \cite{26-CVMH-UNet}  & 85.98 & 75.97 & 85.82 \\
PPMamba \cite{40-PPMamba}      & 91.32 & 84.37 & -     \\
PyramidMamba \cite{42-PyramidMamba} & - & 83.10 & -     \\
PPMamba \cite{56-PPMamba}      & 88.34 & 79.60 & -     \\
UrbanSSF \cite{109-UrbanSSF}    & 91.70 & 85.00 & 93.60 \\ \hline
\end{tabular}%
}

\label{tab:segmentation}
\end{table}

\begin{table}[h]
\caption{The \textbf{change detection} benchmark for \textbf{WHU-CD} dataset \cite{WHUCD}.}
\resizebox{\linewidth}{!}{%
\begin{tabular}{l|ccccc}
\hline
\textbf{Methods} & \textbf{Precision} & \textbf{Recall} & \textbf{F1} & \textbf{IoU} & \textbf{OA} \\ \hline
\multicolumn{6}{c}{\textbf{CNN-based Methods}} \\ \hline
FC-EF \cite{cd_FC_EF} & 92.10 & 90.64 & 91.36 & 84.10 & 99.32 \\
IFNet \cite{cd_IFNet} & 91.51 & 88.01 & 89.73 & 81.37 & 99.20 \\
SNUNet \cite{cd_SNUNet} & 84.70 & 89.73 & 87.14 & 77.22 & 98.95 \\
DSIFN \cite{cd_DSIFN} & \textbf{97.46} & 83.45 & 89.91 & 81.67 & 99.31 \\
CGNet \cite{cd_CGNet} & 94.47 & 90.79 & 92.59 & 86.21 & 99.48 \\ \hline
\multicolumn{6}{c}{\textbf{Transformer-based Methods}} \\ \hline
SwinUNet \cite{cd_SwinUnet} & 92.44 & 87.56 & 89.93 & 81.71 & 99.22 \\
BIT \cite{cd_BIT} & 91.84 & 91.95 & 91.90 & 85.01 & 99.35 \\
ChangeFormer \cite{cd_ChangeFormer} & 93.73 & 87.11 & 90.30 & 82.32 & 99.26 \\
MSCANet \cite{cd_MSCANet}  & 93.47 & 89.16 & 91.27 & 83.94 & 99.32 \\
Paformer \cite{cd_PaFormer}& 94.28 & 90.38 & 92.26 & 85.69 & 99.40 \\ \hline
\multicolumn{6}{c}{\textbf{Mamba-based Methods}} \\ \hline
RS-Mamba \cite{1-RS-Mamba} & 93.37 & 90.42 & 91.87 & 84.96 & - \\
CDMamba \cite{6-CDMamba} & 95.58 & 92.01 & 93.76 & 88.26 & 99.51 \\
IMDCD \cite{8-IMDCD} & 93.85 & 93.27 & 93.56 & 88.39 & 99.51 \\
ChangeMamba \cite{14-ChangeMamba} & 96.18 & 92.23 & 94.19 & 89.02 & \textbf{99.58} \\
DC-Mamba \cite{16-DC-Mamba} & - & 94.33 & \textbf{95.22} & 90.87 & 99.48 \\
TTMGNet \cite{68-TTMGNet} & 92.18 & 89.74 & 90.94 & \textbf{91.25} & 99.15 \\
LCCDMamba \cite{102-LCCDMamba} & 93.41 & \textbf{94.96} & 94.18 & 89.00 & 99.49 \\
CD-Lamba \cite{110-CD-Lamba} & 93.45 & 91.59 & 92.51 & 86.07 & 99.32 \\ \hline
\end{tabular}%
}
\label{tab:changedetection}
\end{table}

\begin{table}[h]
\caption{The \textbf{object detection} benchmark for \textbf{DroneVehicle} \cite{dronevehicle} and \textbf{VisDrone} \cite{Visdrone} datasets. \textbf{mAP} represents mean average precision; \textbf{mAP}$_{0.5}$ represents the mAP at an IoU threshold of 50\%. $\mathcal{M}$ and $\mathcal{Y}$ represent Mamba and YOLO for short. $^{\dagger}$ represents the use of multimodal data, RGB and thermal infrared images.}
\resizebox{\linewidth}{!}{%
\begin{tabular}{lcclc}
\hline
\multicolumn{1}{c|}{\textbf{DroneVehicle \cite{dronevehicle}}} & \textbf{mAP}$_{0.5}$ & \multicolumn{1}{c|}{\textbf{mAP}} & \multicolumn{1}{c|}{\textbf{VisDrone \cite{Visdrone}}} & \multicolumn{1}{l}{\textbf{mAP}} \\ \hline
\multicolumn{5}{c}{\textbf{Transformer/CNN-Based Methods}}                                                                                                                            \\ \hline
\multicolumn{1}{l|}{MKD \cite{od_MKD}}             & -                & \multicolumn{1}{c|}{69.0}         & \multicolumn{1}{l|}{YOLOv10 \cite{od_yolov10}}        & 41.3      \\
\multicolumn{1}{l|}{GHOST \cite{od_ghost}}         & 81.5             & \multicolumn{1}{c|}{59.3}         & \multicolumn{1}{l|}{Gold-YOLO \cite{od_gold}}         & 41.0      \\
\multicolumn{1}{l|}{GM-DETR \cite{od_gmdetr}}      & 80.8             & \multicolumn{1}{c|}{55.9}         & \multicolumn{1}{l|}{DREN \cite{od_DREN}}              & 30.3      \\
\multicolumn{1}{l|}{CMADet \cite{od_CMADet}}       & 82.0             & \multicolumn{1}{c|}{59.5}         & \multicolumn{1}{l|}{GLSAN \cite{od_GLSAN}}         & 32.5      \\ 
\multicolumn{1}{l|}{TSFADet$^{\dagger}$ \cite{od_TSFADet}}     & -                & \multicolumn{1}{c|}{73.9}         & \multicolumn{1}{l|}{TPH-YOLOv5 \cite{od_TPH}}         & 38.9      \\
\multicolumn{1}{l|}{C2Former$^{\dagger}$ \cite{od_C2Forer}}    & -                & \multicolumn{1}{c|}{74.2}         & \multicolumn{1}{l|}{FFCA-YOLO \cite{od_FFCAYOLO}}     & 41.2      \\ \hline
\multicolumn{5}{c}{\textbf{Mamba-Based Methods}}       \\ \hline
\multicolumn{1}{l|}{DMM(R-CNN) \cite{12-DMM}}          & 77.2   & \multicolumn{1}{c|}{-}    & \multicolumn{1}{l|}{ES-HS-FPN \cite{75-ES-HS-FPN}}      & \textbf{43.5}      \\
\multicolumn{1}{l|}{DMM(S2ANet) \cite{12-DMM}}                & 79.4   & \multicolumn{1}{c|}{-}    & \multicolumn{1}{l|}{HRMamba-$\mathcal{Y}$ \cite{94-HRMamba-YOLO}}   & 38.9   \\

\multicolumn{1}{l|}{COMO($\mathcal{Y}$v5s) \cite{84-COMO}}       & 85.3     & \multicolumn{1}{c|}{63.4} & \multicolumn{1}{l|}{Mamba-$\mathcal{Y}$ \cite{94-HRMamba-YOLO}}     & 41.9    \\
\multicolumn{1}{l|}{COMO($\mathcal{Y}$v8s) \cite{84-COMO}}       & \textbf{86.1}     & \multicolumn{1}{c|}{65.5} & \multicolumn{1}{l|}{}                  & \multicolumn{1}{l}{}           \\
\multicolumn{1}{l|}{RemoteDet-$\mathcal{M}$ $^{\dagger}$ \cite{32-RemoteDet-Mamba}}& -      & \multicolumn{1}{c|}{\textbf{81.8}} & \multicolumn{1}{l|}{}     &    \\
\multicolumn{1}{l|}{MGMF$^{\dagger}$ \cite{97-MGMF}}                      & 80.3   & \multicolumn{1}{c|}{-} & \multicolumn{1}{l|}{}                  & \multicolumn{1}{l}{}           \\ \hline
\end{tabular}
}
\label{tab:objectdetection}
\end{table}

\begin{table}[h]
\caption{The \textbf{super-resolution} benchmark for Chikusei \cite{sr_Chi} and AID \cite{sr_AID} datasets (scale factor is 4 $\times$). \textbf{PSNR} represents peak signal-to-noise ratio and \textbf{SSIM} represents structural similarity index. The data on Chikusei \cite{sr_Chi} are preprocessed differently so that the performances of HSRMamba \cite{111-HSRMamba} and MambaIR \cite{sr_MambaIR} are better.}
\resizebox{\linewidth}{!}{ 
\begin{tabular}{lcclcc}
\hline
\multicolumn{1}{c|}{\multirow{1}{*}{\textbf{Chikusei \cite{sr_Chi}}}}    & \textbf{PSNR}             & \multicolumn{1}{c|}{\textbf{SSIM}} & \multicolumn{1}{c|}{\multirow{1}{*}{\textbf{AID \cite{sr_AID}}}} & \textbf{PSNR}        & \textbf{SSIM}        \\ \hline
\multicolumn{6}{c}{\textbf{CNN-based Methods}}  \\ \hline
\multicolumn{1}{l|}{SSR-NET \cite{sr_SSRNet}}                            & 25.37                     & \multicolumn{1}{c|}{-}   & \multicolumn{1}{l|}{FENet \cite{sr_FeNet}} & 29.16 & 0.7812 \\ \hline
\multicolumn{6}{c}{\textbf{Transformer-based Methods}} \\ \hline
\multicolumn{1}{l|}{{\color[HTML]{212529} MSST-Net \cite{sr_MMSTNet}}} & 22.66                     & \multicolumn{1}{c|}{-}             & \multicolumn{1}{l|}{HAT-L \cite{sr_HAT}}                         & 30.81                & 0.8124               \\
\multicolumn{1}{l|}{{\color[HTML]{212529} UHNTC \cite{sr_UHNTC}}}    & \multicolumn{1}{l}{25.76} & \multicolumn{1}{c|}{-}  & \multicolumn{1}{l|}{RGT  \cite{sr_RGT}} & 30.91  & 0.8159  \\
\multicolumn{1}{l|}{{\color[HTML]{212529} MIMO–SST \cite{sr_MIMOSST}}}   & \multicolumn{1}{l}{28.78} & \multicolumn{1}{c|}{-}  & \multicolumn{1}{l|}{TransENet \cite{sr_TransUNet}}  & 30.80   & 0.8109     \\ \hline
\multicolumn{6}{c}{\textbf{Mamba-based Methods}}  \\ \hline
\multicolumn{1}{l|}{UVMSR \cite{44-UVMSR}} & 28.12 & \multicolumn{1}{c|}{\textbf{0.9642}} & \multicolumn{1}{l|}{FreMamba \cite{4-FreMamba}}   & \textbf{31.07}    & \textbf{0.8185}               \\
\multicolumn{1}{l|}{SSRFN \cite{52-SSRFN}}    & \textbf{29.86}  & \multicolumn{1}{c|}{-}  & \multicolumn{1}{l|}{MambaFormerSR \cite{34-MambaFormer}} & 29.35 & 0.7870   \\
\multicolumn{1}{l|}{FusionMamba \cite{2-RS3Mamba}} & 27.61   & \multicolumn{1}{c|}{-}  & \multicolumn{1}{l|}{MambaIR \cite{sr_MambaIR}} & 30.85  & 0.8130   \\
\multicolumn{1}{l|}{HSRMamba \cite{111-HSRMamba}}                          & \textbf{40.28}                     & \multicolumn{1}{c|}{0.9441}        & \multicolumn{1}{l|}{}                                   & \multicolumn{1}{l}{} & \multicolumn{1}{l}{} \\
\multicolumn{1}{l|}{MambaIR \cite{sr_MambaIR}}                               & 39.48                     & \multicolumn{1}{c|}{0.9353}        & \multicolumn{1}{l|}{}                                   & \multicolumn{1}{l}{} & \multicolumn{1}{l}{} \\ \hline
\end{tabular}
}
\label{tab:superresolution}
\end{table}

\begin{table}[h]
\caption{The \textbf{image restoration} benchmark for SateHaze1k \cite{SateHaze1k} and UAV-Rain \cite{UAVRain}.}
\resizebox{\linewidth}{!}{ 
\begin{tabular}{llllll}
\hline
\multicolumn{1}{c|}{}   & \multicolumn{2}{c|}{\textbf{SateHaze1k}} & \multicolumn{1}{c|}{}   & \multicolumn{2}{c}{\textbf{UAV-Rain}}   \\ \cline{2-3} \cline{5-6} 
\multicolumn{1}{c|}{\multirow{-2}{*}{\textbf{Methods}}} & \multicolumn{1}{c}{\textbf{PSNR}} & \multicolumn{1}{c|}{\textbf{SSIM}} & \multicolumn{1}{c|}{\multirow{-2}{*}{\textbf{Methods}}} & \multicolumn{1}{c}{\textbf{PSNR}} & \multicolumn{1}{c}{\textbf{SSIM}} \\ \hline
\multicolumn{6}{c}{\textbf{CNN-based Methods}}  \\ \hline
\multicolumn{1}{l|}{FFA-Net \cite{ir_FFANet}}    & 22.87  & \multicolumn{1}{l|}{0.8965} & \multicolumn{1}{l|}{RCDNet \cite{ir_RCDNet}} & 22.48   & 0.8753        \\
\multicolumn{1}{l|}{M2SCN \cite{ir_M2SCN}}      & 24.22  & \multicolumn{1}{l|}{0.8960} & \multicolumn{1}{l|}{SPDNet \cite{ir_SPDNet}} & 24.78   & 0.9054        \\ \hline
\multicolumn{6}{c}{\textbf{Transformer-based Methods}}  \\ \hline
\multicolumn{1}{l|}{Restormer \cite{ir_Restormer}}  & 24.34  & \multicolumn{1}{l|}{0.9021}  & \multicolumn{1}{l|}{IDT \cite{ir_IDT}}   & 22.47   & 0.9054                            \\
\multicolumn{1}{l|}{RSDformer \cite{ir_RSDformer}}  & 24.30  & \multicolumn{1}{l|}{0.9071}  & \multicolumn{1}{l|}{DRSformer \cite{ir_DRSformer}}  & 24.93   & \textbf{0.9155}                            \\ \hline
\multicolumn{6}{c}{\textbf{Mamba-based Methods}}  \\ \hline
\multicolumn{1}{l|}{RSDehamba \cite{7-RSDehamba}}   & \textbf{25.91}    & \multicolumn{1}{l|}{\textbf{0.9157}}   & \multicolumn{1}{l|}{LightMamba \cite{24-LightMamba}}     & \textbf{25.56}      & 0.9042                            \\
\multicolumn{1}{l|}{LightMamba \cite{24-LightMamba}}     & 25.80    & \multicolumn{1}{l|}{0.9148}   & \multicolumn{1}{l|}{Weamba \cite{106-Weaba}}          & 25.25    & 0.9080       \\
\multicolumn{1}{l|}{FMambaIR \cite{80-FMambaIR}}   & 24.35  & \multicolumn{1}{l|}{0.9003}    & \multicolumn{1}{l|}{}   &        &          \\
\multicolumn{1}{l|}{MambaIR \cite{sr_MambaIR}}    & 24.50        & \multicolumn{1}{l|}{0.9093}  & \multicolumn{1}{l|}{}        &           &               \\
\multicolumn{1}{l|}{Weamba \cite{106-Weaba}}  & 25.55  & \multicolumn{1}{l|}{0.9151}  & \multicolumn{1}{l|}{}    &         &                    \\ \hline
\end{tabular}
}
\label{tab:restoration}
\end{table}


In this section, we introduce six benchmarks involving models based on CNN, Transformer, and Mamba architectures, across various downstream tasks. Tab. \ref{tab:summarypaper} lists five main tasks in remote sensing and their representative works.
To provide a more comprehensive comparison and demonstration, we have selected tasks that include both high-level vision tasks (i.e., image classification (Tab.~\ref{tab:classification}), segmentation (Tab.~\ref{tab:segmentation}), change detection (Tab.~\ref{tab:changedetection}, object detection (Tab.~\ref{tab:objectdetection}) and low-level vision tasks (i.e., super-resolution (Tab.~\ref{tab:superresolution}), image restoration (Tab.~\ref{tab:restoration})). 

\noindent \par \textbf{Results on Image Classification Task:} Tab.~\ref{tab:classification} shows the classification performance among different models on the Pavia University Dataset \cite{pu}. As we can see, although the Rate parameters set by each method vary, when the same Rate (5.00\%) is adopted, the IGroupSS-Mamba \cite{51-IGroupSS-Mamba} achieves optimal performance in parameters such as OA, AA, Kappa, and surpasses all other CNN-based and Transformer-based models.
\noindent \par \textbf{Results on Image Segmentation Task:} Tab.~\ref{tab:segmentation} presents the comparison performance among different models on the ISPRS Vaihingen dataset~\cite{vaihingen}. Overall, Transformer-based methods tend to outperform CNN-based methods, but Mamba-based methods generally surpass Transformer-based methods in most cases. Specifically, PPMamba \cite{40-PPMamba} achieves the best performance on the mF1 and mIOU metrics, while \cite{25-CM-UNet} excels in the OA metric among all methods reporting this metric.
\noindent \par \textbf{Results on Image Change Detection Task:} Tab.~\ref{tab:changedetection} shows the performance comparisons among different models on the WHU-CD dataset~\cite{WHUCD}. Similarly, Mamba-based models achieve better performance than CNN-based and Transformer-based models. Specifically, DC-Mamba \cite{16-DC-Mamba} delivers the best performance in the F1 metric, TTMGNet \cite{68-TTMGNet} excels in the IoU metric, and ChangeFormer \cite{cd_ChangeFormer} achieves the best performance in the OA metric. For more comprehensive comparison, it is recommended that the author refer to the relevant survey paper~\cite{DBLP:journals/remotesensing/ChengHLLXZZX24}.
\noindent \par \textbf{Results on Image Object Detection Task:} Tab.~\ref{tab:objectdetection} compares the performance of object detection tasks on the DroneVehicle~\cite{dronevehicle} and VisDrone~\cite{Visdrone} datasets. Specifically, the Mamba-based model, ES-HS-FPN \cite{75-ES-HS-FPN}, outperforms both CNN-based and Transformer-based models, achieving the highest mAP value.
\noindent \par \textbf{Results on Image Super-Resolution Task:} Table~\ref{tab:superresolution} presents a comparison of image super-resolution task performance across the Chikusei \cite{sr_Chi} and AID \cite{sr_AID} datasets. In particular, FreMamba \cite{4-FreMamba} records the highest values in both PSNR and SSIM metrics on the AID dataset.
\noindent \par \textbf{Results on Image Restoration Task:} Tab.~\ref{tab:restoration} presents a performance comparison of image restoration tasks on the SateHaze1k \cite{SateHaze1k} and UAV-Rain \cite{UAVRain} datasets. Notably, RSDehamba \cite{7-RSDehamba} sets a new state-of-the-art benchmark for the tasks on the SateHaze1k dataset in both PSNR and SSIM metrics.

%% file: Sec/7_future.tex
\section{Challenges and Future Directions} \label{sec:direction}
This section examines the principal challenges encountered when applying Mamba to remote sensing and outlines several promising directions for future research to enhance its performance in this domain.

\subsection{Causality} 
The S6 block in Mamba operates as a causal system \cite{mamba}, wherein predictions for each token are generated based solely on the current input and a hidden state summarizing preceding information. This architecture is inherently optimized for one-dimensional sequential data rather than two-dimensional images. Although patch embedding techniques and flattening allow to transform 2D data into 1D sequences, this process inevitably results in the loss of spatial information due to the causal nature of the system, thereby compromising performance \cite{survey1, survey3, survey4, survey5, survey6, surveyon2 }.

One strategy to alleviate this issue involves employing more effective scanning strategies. As detailed in Section \ref{sec:scanstrategy}, several scan methods have been proposed to partially overcome causal constraints and reduce spatial information loss. Nevertheless, these methods do not entirely resolve the fundamental limitations. An alternative approach is to modify the core formulation of the SSM itself. For instance, VSSD \cite{VSSD} incorporates a backbone architecture specifically designed for non-causal data processing, whereas TTMGNet \cite{68-TTMGNet} adjusts its hidden state calculations to match the employed scanning strategy, unintentionally shifting towards a non-causal framework. Developing an SSM explicitly for handling non-causal data remains an attractive and relatively unexplored area of research.


\subsection{Novel SSM Formulations} 
As discussed in Section \ref{sec:SSM}, several studies have proposed modifications to the SSM formulation to achieve specific objectives. Nevertheless, this research area remains in its early stages, with considerable scope for further improvement. Developing an SSM formulation that is optimally suited to the characteristics of remote sensing imagery is both critical and promising for future studies.

\subsection{Multi-Modal Interaction via Mamba} 
Employing multi-modal and bi-temporal data in remote sensing is prevalent. While certain studies discussed in Section \ref{sec:multimodality} have explored multimodal and bi-temporal data interaction using Mamba, further efforts are necessary to develop more efficient and interpretable interaction architectures based on Mamba. Advancements in this direction could significantly enhance the performance of multimodal and bi-temporal remote sensing tasks.


\subsection{3D Scan Data Processing} 
For data rich in spectral information, such as the hyperspectral image, an intriguing approach is to consider these data as 3D data. Within the Mamba framework, only SSUMamba \cite{79-SSUMamba} currently leverages a 3D scan strategy to process such data. Further exploration of 3D scanning techniques to capture spatial-spectral relationships represents an unexplored but promising direction in remote sensing research.

\subsection{Mamba-based Foundation Models in Remote Sensing} 
The recent success of foundation models \cite{foundationsurvey} across various domains has spurred interest in their application to remote sensing. However, to date, only SatMAE \cite{SatMAE} has provided a preliminary validation of Mamba-based foundation models in this field. Given Mamba’s theoretical computational efficiency and linear complexity, Mamba-based models may offer significant advantages over transformer-based models, particularly when processing very high-resolution images.

Nonetheless, scaling Mamba-based models to larger sizes may introduce stability issues \cite{survey4}. Although several backbone networks \cite{ARM, SiMBA, StableMamba, Mambav2} have proposed methods to enhance stability in large-scale implementations, these solutions have not yet been fully validated for foundation models in remote sensing. Addressing the stability challenges associated with large Mamba-based models is thus a crucial area for future research.

\subsection{Computational Efficiency} 
Despite the advantages of linear computational complexity and a low computation burden, Mamba’s recurrent computation paradigm leads to relatively low computational efficiency. The Mamba-1 \cite{mamba} incorporates a hardware-aware algorithm for acceleration, and Mamba-2 \cite{Mambav2} further improves efficiency by introducing an optimized SSD algorithm, achieving a 2–8$\times$ speedup compared to the vanilla. However, many remote sensing applications utilizing Mamba-1 still require enhancements in computational efficiency. Potential approaches include: (1) adopting Mamba-2 for remote sensing tasks, (2) developing more efficient hardware-aware algorithms, and (3) modifying the SSM formulation to circumvent the inherent limitations of recurrent computation, thereby enabling more effective utilization of current GPUs.

\subsection{Adaptations to Downstream Tasks} 
While Mamba-based models have demonstrated promise in several remote sensing applications, their full potential remains underexplored, particularly in unconventional or emerging downstream tasks. 
The architecture’s advantages, such as linear scaling and efficient long-range dependency modeling, could effectively address unique challenges in less-studied tasks, such as some application in agriculture \cite{survey_agri}, in forest \cite{survey_for} and in ecological restoration \cite{survey_eco} and in less-explored data, such as hyperspectral time-series analysis \cite{TSH} that is characterized by both high spectral resolution and temporal continuity. 
Moreover, the potential of vision-language (CV + NLP) for remote sensing remains underexplored. A Mamba-based multimodal framework could enable novel applications such as content-guided image retrieval \cite{image_retrive_survey} in remote sensing data, automated visual question answering (VQA) \cite{VQA_Survey} for non-expert users and image captioning \cite{imageCaptioning} of remote sensing imagery.
In addition, given the recent success of Large Language Models (LLMs), combining Vision Mamba with LLMs represents a promising direction, as it leverages the strong reasoning capabilities of LLMs.
Beyond the tasks discussed in Section \ref{sec:downstramtask}, further investigation into applications traditionally addressed by CNN and Transformer models represents a significant and promising research direction.